\documentclass{article}

\PassOptionsToPackage{numbers, compress, sort}{natbib}
\newcommand{\FedAvg}{\method{FedAvg}\xspace}

\newcommand{\FedProx}{\method{FedProx}\xspace}
\newcommand{\SCAFFOLD}{\method{SCAFFOLD}\xspace}
\newcommand{\FedDANE}{\method{FedDANE}\xspace}

\newcommand{\SGD}{\method{SGD}\xspace}

\newcommand{\FedRoD}{\method{Fed-RoD}\xspace}
\newcommand{\Mime}{\method{Mime}\xspace}
\newcommand{\FedDyn}{\method{FedDyn}\xspace}
\newcommand{\MOCHA}{\method{MTL}\xspace}
\newcommand{\LGFedAvg}{\method{LG-FedAvg}\xspace}
\newcommand{\pFedMe}{\method{pFedMe}\xspace}
\newcommand{\PerFedAvg}{\method{Per-FedAvg}\xspace}
\newcommand{\FedFOMO}{\method{FedFOMO}\xspace}
\newcommand{\FMTL}{\method{Ditto}\xspace}
\newcommand{\FedPer}{\method{FedPer}\xspace}
\newcommand{\pFedHN}{\method{pFedHN}\xspace}
\newcommand{\FedRep}{\method{FedRep}\xspace}
\usepackage{amssymb}
\usepackage{amsmath,amsfonts}
\usepackage{mathtools}
\usepackage{amsopn}
\usepackage{bm} 
\usepackage{multirow}

\newcommand{\vct}[1]{\boldsymbol{#1}} 
\newcommand{\mat}[1]{\boldsymbol{#1}} 

\newcommand{\field}[1]{\mathbb{#1}}
\newcommand{\R}{\field{R}} 
\newcommand{\C}{\field{C}} 

\newcommand{\ProbOpr}[1]{\mathbb{#1}}

\newcommand{\expect}[2]{%
\ifthenelse{\equal{#2}{}}{\ProbOpr{E}_{#1}}
{\ifthenelse{\equal{#1}{}}{\ProbOpr{E}\left[#2\right]}{\ProbOpr{E}_{#1}\left[#2\right]}}} 

\DeclareMathOperator{\argmax}{arg\,max}
\DeclareMathOperator{\argmin}{arg\,min}

\newcommand{\vtheta}{\vct{\theta}}

\newcommand{\vzeta}{\vct{\zeta}}

\newcommand{\vq}{\vct{q}}
\newcommand{\vx}{{\vct{x}}}

\newcommand{\vz}{{\vct{z}}}
\newcommand{\vnu}{\vct{\nu}}

\newcommand{\vw}{\vct{w}}

\newcommand{\vphi}{\vct{\phi}}
\newcommand{\vpsi}{\vct{\psi}}

\newcommand{\mOmega}{\mat{\Omega}}

\newcommand{\sD}{\mathcal{D}}
\newcommand{\sL}{\mathcal{L}}
\newcommand{\sP}{\mathcal{P}}
\newcommand{\sR}{\mathcal{R}}
\newcommand{\eat}[1]{}
\newcommand{\method}[1]{\textsc{#1}}

\usepackage[dvipsnames]{xcolor}
\usepackage{iclr2022_conference,times}
\usepackage{wrapfig}
\usepackage[utf8]{inputenc} 
\usepackage[pagebackref=true, colorlinks=true, allcolors=red, citecolor=green]{hyperref}       
\usepackage[T1]{fontenc}    
\usepackage{hyperref}       
\usepackage{url}            
\usepackage{booktabs}       
\usepackage{amsfonts}       
\usepackage{nicefrac}       
\usepackage{microtype}      

\usepackage{hyperref}
\usepackage{graphicx}
\usepackage{subfigure}
\usepackage{url}            
\usepackage{booktabs}       
\usepackage{amsfonts}       
\usepackage{nicefrac}       
\usepackage{enumitem}

\usepackage{wrapfig}
\usepackage{xspace}
\usepackage{caption}
\usepackage[ruled, vlined]{algorithm2e}

\usepackage{physics}
\usepackage{lipsum}
\usepackage{array}
\usepackage{boldline}
\usepackage{enumitem}

\usepackage{colortbl}
\definecolor{Gray}{gray}{0.9}
\definecolor{LightCyan}{rgb}{0.88,1,1}
\renewcommand{\paragraph}[1]{\vspace{-0.5ex}\textbf{#1}}

\newcommand{\eg}{{\em e.g.}}
\newcommand{\ie}{{\em i.e.}}

\newsavebox\mybox
\newlength\mylength

\iclrfinalcopy

\title{On Bridging Generic and Personalized\\ Federated Learning for Image Classification}

\author{%
  Hong-You Chen \\
  The Ohio State University, USA\\
  \And
  Wei-Lun Chao \\
  The Ohio State University, USA\\

}

\begin{document}

\maketitle


\begin{abstract}
Federated learning is promising for its capability to collaboratively train models with multiple clients without accessing their data, but vulnerable when clients' data distributions diverge from each other. 
This divergence further leads to a dilemma: \emph{``Should we prioritize the learned model's generic performance (for future use at the server) or its personalized performance (for each client)?''} These two, seemingly competing goals have divided the community to focus on one or the other, yet in this paper we show that it is possible to approach both at the same time.
Concretely, we propose a novel federated learning framework that explicitly \emph{decouples a model's dual duties with two prediction tasks}.
On the one hand, we introduce a family of losses that are robust to non-identical class distributions, enabling clients to train a \emph{generic} predictor with a consistent objective across them. 
On the other hand, we formulate the \emph{personalized} predictor as a lightweight adaptive module that is learned to minimize each client's empirical risk on top of the generic predictor.
With this \emph{two-loss, two-predictor} framework which we name \textbf{Federated Robust Decoupling (\FedRoD)}, the learned model can simultaneously achieve state-of-the-art generic and personalized performance, essentially bridging the two tasks. 
\end{abstract}

\section{Introduction}
\label{s_intro}

Large-scale data are the driving forces for modern machine learning but come with the risk of data privacy. In applications like health care, data are required to be kept separate to enforce ownership and protection, hindering the collective wisdom (of data) for training strong models. Federated learning (FL), 
which aims to train a model with multiple data sources (\ie, clients)
while keeping their data decentralized, has emerged as a popular paradigm to resolve these concerns~\citep{kairouz2019advances}.

\begin{wrapfigure}{r}{0.45\columnwidth}
    \centering
    {\includegraphics[width=0.44\columnwidth]{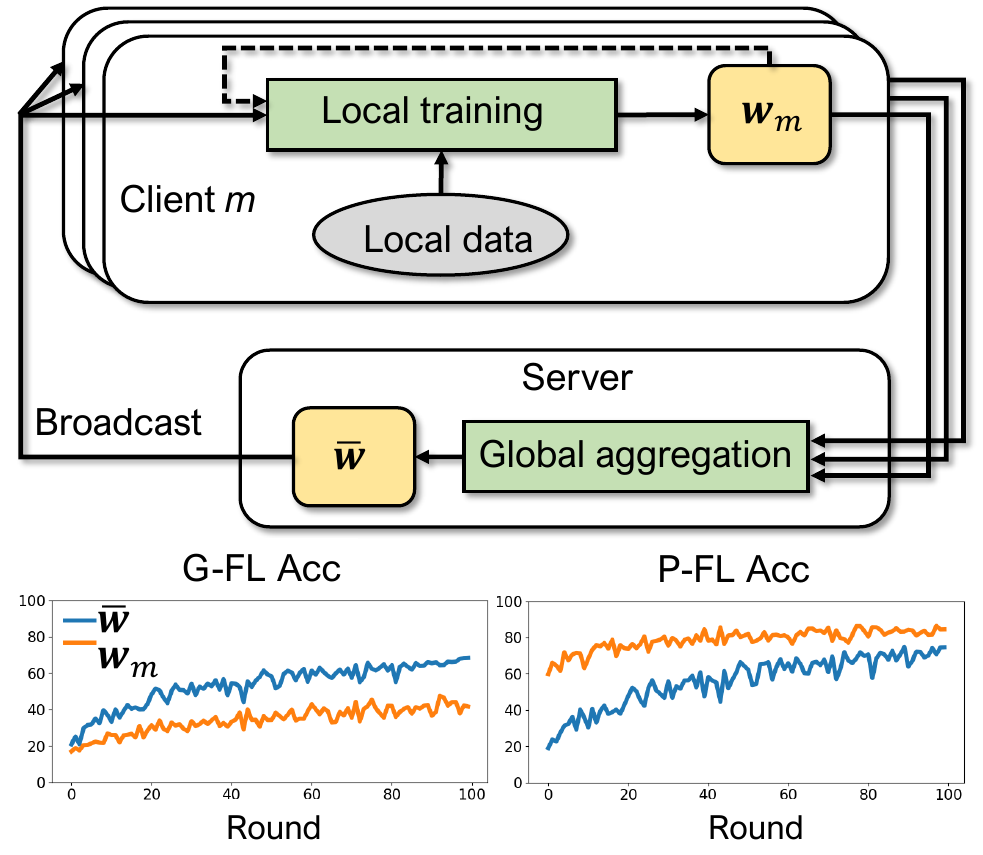}}
    \vskip -13pt
    \caption{\small \textbf{The multi-round generic FL pipeline (top).} The dashed arrow indicates that local models or statistics may be carried to the next round. Here we apply \FedAvg \citep{mcmahan2017communication} on CIFAR-10 with 20 non-IID clients (see \autoref{s_exp}), and show that \emph{personalized models come for free from generic FL} \textbf{(bottom)}.
    The global model {\color{RoyalBlue}$\bar{\vw}$} outperforms local models {\color{RedOrange}$\vw_m$} on the bottom-left generic accuracy (G-FL), yet {\color{RedOrange}$\vw_m$} outperforms {\color{RoyalBlue}$\bar{\vw}$} on the bottom-right personalized accuracy (P-FL). The accuracy is computed at the end of each round.
    } \label{fig:overview}
    \vskip -11pt
\end{wrapfigure}

The standard setup of FL seeks to train a single ``global'' model that can perform well on \emph{generic} data distributions \citep{kairouz2019advances}, \eg, the union of clients' data.
As clients' data are kept separate,
mainstream algorithms like \FedAvg~\citep{mcmahan2017communication} take a multi-round approach shown in \autoref{fig:overview}. Within each round, the server first broadcasts the ``global'' model to the clients, who then independently update the model locally 
using their own (often limited) data. 
The server then aggregates the ``local'' models back into the ``global'' model 
and proceeds to the next round. 
This pipeline is shown promising if {clients' data are IID} (\ie, with similar data and label distributions)~\citep{zhou2017convergence,stich2019local}, which is, however, hard to meet in reality and thus results in a drastic performance drop~\citep{li2020convergence,zhao2018federated}.
Instead of sticking to a single ``global'' model that features the generic performance, another setup of FL seeks to construct a ``personalized'' model for each client to acknowledge the heterogeneity among clients 
\citep{smith2017federated,dinh2020personalized,hanzely2020lower}. 
This latter setup (usually called \textbf{personalized FL}) is shown to outperform the former (which we name \textbf{generic FL}) regarding the test accuracy of each client alone.

\emph{So far, these two seemingly contrasting FL setups are developed independently. In this paper, we however found that they can be approached simultaneously by generic FL algorithms like \FedAvg.}

Concretely, algorithms designed for generic FL (G-FL) often discard the local models $\{\vw_m\}$ after training (see \autoref{fig:overview}). As a result, when they are evaluated in a personalized setting (P-FL), it is the global model $\bar{\vw}$ being tested~\citep{fallah2020personalized,arivazhagan2019federated,liang2020think,fedfomo,dinh2020personalized,smith2017federated,li2020federated-FMTL}. Here, we found that if we instead keep $\{\vw_m\}$ and evaluate them in P-FL, they outperform nearly all the existing P-FL algorithms. In other words, \emph{personalized models seem to come for free from the local training step of generic FL}.   

At first glance, this may not be totally surprising: local training in G-FL algorithms is driven by the client's empirical risk, which is what a personalized model strives to optimize\footnote{However, when G-FL algorithms are tested on the P-FL setup,
the literature does not use their local models.}.
What really surprises us is that even without an explicit regularization term imposed by most P-FL algorithms \citep{smith2017federated,dinh2020personalized}, the local models of G-FL algorithms can achieve better generalization performance. We conduct a detailed analysis and argue that global aggregation --- taking average over model weights --- indeed acts like a regularizer for local models. 
Moreover, applying advanced G-FL algorithms~\citep{feddyn} to improve the G-FL accuracy seems to not hurt the  ``local'' models' P-FL accuracy.

Building upon these observations, we dig deeper into generic FL.
Specifically for classification, the non-IID clients can result from non-identical \textbf{class distributions} or non-identical \textbf{class-conditional data distributions}. One way to mitigate their influences is to make the local training objectives more aligned among clients.
While this can hardly be achieved for the latter case without knowing clients' data, we can do so for the former case by setting a consistent goal among clients --- \emph{the learned local models should classify every class well, even if clients' data have different class distributions.}
We realize this by viewing each client's local training as an independent class-imbalanced problem~\citep{cui2019class,he2009learning}
and applying objective functions dedicated to it \citep{cao2019learning,ren2020balanced}. 
As will be shown in \autoref{s_exp}, these class-balanced objectives lead to much consistent local training among clients, making the resulting ``global'' model more robust to non-IID conditions. 

The use of class-balanced objectives, nevertheless, degrades the local models' P-FL performance. This is because the local models are no longer learned to optimize clients' empirical risks. 

\begin{wrapfigure}{r}{0.45\columnwidth}
\vskip -16pt
    \centering
    {\includegraphics[width=0.45\columnwidth]{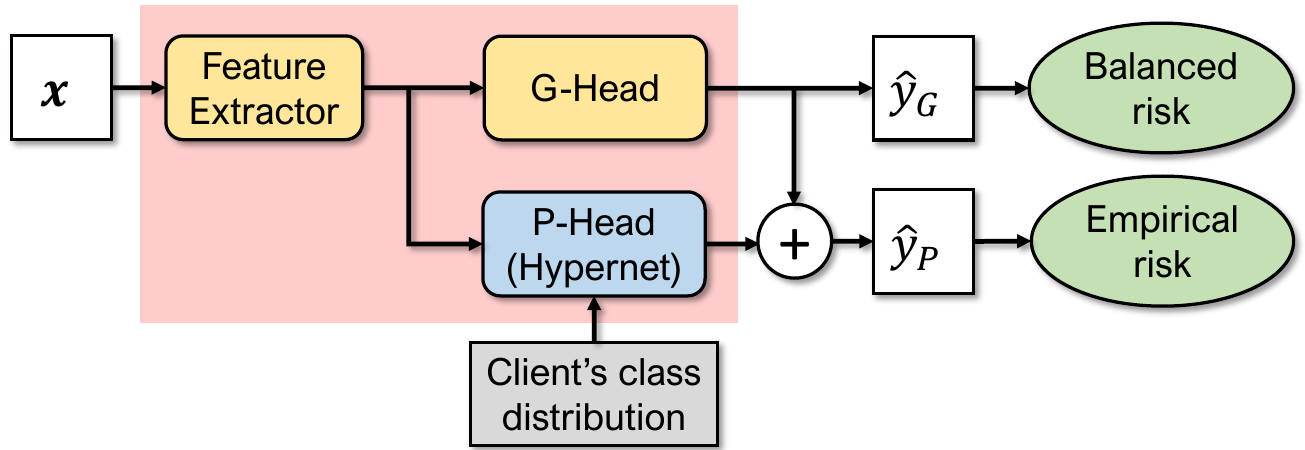}}%
    \vskip -8pt
    \caption{\small \textbf{Local training of \FedRoD.} {\color{Dandelion}Yellow}/{\color{RoyalBlue}blue} boxes are the models for G-FL/P-FL. {\color{LimeGreen}Green} ellipsoids are the learning objectives. The {\color{CarnationPink}red} area means what to be aggregated at the server.} \label{fig:fedrod}
    \vskip -10pt
\end{wrapfigure}

To address this issue, we propose a unifying framework for G-FL and P-FL which explicitly \emph{decouples a local model's dual duties}: serving as the personalized model and the ingredient of the global model.
Concretely, we follow the \FedAvg pipeline and train the local model with the \emph{class-balanced loss}, but on top of the feature extractor, we introduce a lightweight personalized predictor and train it with client's \emph{empirical risk} (see \autoref{fig:fedrod}). With this \emph{two-loss, two-predictor} framework which we name \textbf{Federated Robust Decoupling (\FedRoD)}, the resulting global model can be more robust to non-identical class distributions; the personalized predictor can lead to decent P-FL accuracy due to the implicit regularization and the empirical loss. Specifically for the personalized predictor, we propose to explicitly parameterize it with clients' class distributions via a hypernetwork \citep{ha2016hypernetworks}. That is, we learn a shared meta-model that outputs personalized predictors for clients given their class distributions. This not only enables \emph{zero-shot model adaptation} to new clients (without their data but class distributions), but also provides a better initialization to fine-tune the models given new clients' data.

We validate \FedRoD on multiple datasets under various non-IID settings.
\FedRoD consistently outperforms existing generic and personalized FL algorithms in both setups.
Moreover, \FedRoD is compatible with and can further improve advanced generic FL algorithms like \FedDyn~\citep{feddyn} whenever non-identical class distributions occur. Our contributions are three-folded:
\begin{itemize}[noitemsep,topsep=0pt,parsep=0pt,partopsep=0pt,leftmargin=*]
\item Unlike most of the previous works that focus on either generic FL or personalized FL, we propose \FedRoD to excel on both at the same time. We validate \FedRoD with extensive experiments.
\item We show that strong personalized models emerge from the local training step of generic FL algorithms, due to implicit regularization. We further show that class-balanced objectives are effective for improving the generic FL performance when clients have different class distributions.
\item \FedRoD enables zero-shot adaptation and
much effective fine-tuning for new clients.
\end{itemize} 


\section{Related Work (A detailed version is in \autoref{suppl-sec:related})}
\label{s_related}

\noindent\textbf{Generic federated learning.}
\FedAvg~\citep{mcmahan2017communication} is the standard algorithm, and many works are proposed to improve it, either in the \textbf{global aggregation} step~\citep{yurochkin2019bayesian,wang2020federated,lin2020ensemble,chen2021fedbe,hsu2019measuring,reddi2021adaptive} or \textbf{local training} step~\citep{malinovskiy2020local,yuan2020federated,zhao2018federated,wang2020tackling}. For example, to reduce local models' drifts from the global model, \FedProx \citep{li2020federated} and \FedDyn \citep{feddyn} employed regularization toward the global model; \SCAFFOLD~\citep{karimireddy2020mime} leveraged control variates to correct local gradients. We also aim to reduce local models' drifts but via a different way. We apply objective functions in class-imbalanced learning \citep{he2009learning}, which are designed to be robust to class distribution changes. The closest to ours is \citep{hsu2020federated}, which used a traditional class-imbalanced treatment named re-weighting. We show that more advanced techniques can be applied to further improve the performance, especially under extreme non-IID conditions where re-weighting is ineffective.

\noindent\textbf{Personalized federated learning.} Many approaches for personalized FL \citep{Kulkarni2020SurveyOP} are based on multi-task learning (MTL)~\citep{zhang2017survey,ruder2017overview}. For instance, \cite{smith2017federated} encouraged related clients to learn similar models; \cite{li2020federated-FMTL,dinh2020personalized,hanzely2020lower} regularized local models with a learnable global model. Our approach is inspired by MTL as well but has notable differences. First, we found that global aggregation in generic FL already serves as a strong regularizer. Second, instead of learning for each client a feature extractor \citep{bui2019federated,liang2020think} or an entire model, \FedRoD shares a single feature extractor among clients, inspired by \cite{zhang2014facial,caruana1997multitask}. This reduces the total parameters to be learned and improves generalization. Compared to \citep{arivazhagan2019federated,collins2021exploiting} which also learned a shared feature extractor, \FedRoD simultaneously excels in both FL setups.

Instead of designing specific algorithms for personalized FL,  \cite{yu2020salvaging,Wang2019FederatedEO,cheng2021fine} showed that performing post-processing (\eg, fine-tuning) to a generic FL model (\eg, $\bar{\vw}$ in \FedAvg) leads to promising personalized accuracy. We further showed that, the \emph{local models} $\{\vw_m\}$ learned in \FedAvg and other generic FL algorithms are strong personalized models.

\emph{We note that, while many personalized FL algorithms also produce a global model, it is mainly used to regularize or construct personalized models but not for evaluation in the generic setup. In contrast, we learn models to excel in both setups via a single framework without sacrificing
either of them.}

A recent work \pFedHN~\citep{shamsian2021personalized} also applies hypernetworks~\citep{ha2016hypernetworks} but in a very different way from \FedRoD. \pFedHN learns a hypernetwork at the server to aggregate clients' updates and produce entire models for them for the next round. In contrast, we learn the hypernetwork locally to construct the personalized predictors, not the entire models, for fast adaptation to clients.


\section{Personalized Models Emerge from Generic Federated Learning}
\label{s_GFLPFL}

In this section, we show that personalized FL (P-FL) models emerge from the training process of generic FL (G-FL) algorithms. To begin with, we review representative G-FL and P-FL algorithms.

\subsection{Background}
\label{ss_back_G}
\noindent\textbf{Generic federated learning.} In a generic FL setting with $M$ clients, where each client has a data set $\sD_m = \{(\vx_i, y_i)\}_{i=1}^{|\sD_m|}$, the optimization problem to solve can be formulated as
\begin{align}
\min_{\vw}~\sL(\vw) = \sum_{m=1}^M \frac{|\sD_m|}{|\sD|} \sL_m(\vw), \hspace{10pt}\text{where} \hspace{10pt} \sL_m(\vw) = \frac{1}{|\sD_m|} \sum_{i} \ell(\vx_i, y_i; \vw).
\label{eq:obj}
\end{align}
Here, ${\vw}$ is the model parameter; $\sD = \cup_m \sD_m$ is the aggregated data set from all clients; $\sL_m(\vw)$ is the empirical risk computed from client $m$'s data; $\ell$ is a loss function applied to each data instance.

\noindent\textbf{Federated averaging (\FedAvg).} As clients' data are separate, \autoref{eq:obj} cannot be solved directly. A standard way to \emph{relax} it is \FedAvg \citep{mcmahan2017communication}, which iterates between two steps, local training and global aggregation, for \emph{multiple rounds of communication}
\begin{align}
\textbf{Local: } \hspace{4pt} \vw_m = \argmin_{\vw} \sL_m(\vw), \text{ initialized with } \bar{\vw}; \hspace{20pt} \textbf{Global: } \hspace{4pt} \bar{\vw} \leftarrow \sum_{m=1}^M \frac{|\sD_m|}{|\sD|}{\vw_m}.
\label{eq_gfl_local}
\end{align}
The local training is performed at all (or part of) the clients in parallel, usually with multiple epochs of SGD to produce the local model $\vw_m$.
The global aggregation is by taking element-wise average over model weights. Since local training is driven by clients' empirical risks, when clients' data are non-IID, $\vw_m$ would drift away from each other, making $\bar{\vw}$ deviate from the solution of \autoref{eq:obj}.

\noindent\textbf{Personalized federated learning.}
Personalized FL learns for each client $m$ a model $\vw_m$, whose goal is to perform well on client $m$'s data. While there is no agreed objective function so far, many existing works~\citep{smith2017federated,li2020federated-FMTL,dinh2020personalized,hanzely2020lower,hanzely2020federated,li2019fedmd} define the optimization problems similar to the following
\begin{align}
\min_{\{\mOmega, \vw_1,\cdots,\vw_M\}} \sum_{m=1}^M  \frac{|\sD_m|}{|\sD|}\sL_m(\vw_m) + \sR(\mOmega,\vw_1,\cdots,\vw_M),
\label{eq:P-obj} 
\end{align}
where $\sR$ is a regularizer; $\mOmega$ is introduced to relate clients. The regularizer is imposed to prevent $\vw_m$ from over-fitting client $m$'s limited data.
Unlike \autoref{eq:obj}, \autoref{eq:P-obj} directly seeks to minimize each client's empirical risk (plus a regularization term) by the corresponding personalized model $\vw_m$. 

In practice, personalized FL algorithms often run iteratively between the local and global steps as well, so as to update $\mOmega$ according to clients' models. One example is to define $\mOmega$ as a global model \citep{hanzely2020federated,hanzely2020lower,dinh2020personalized,li2020federated-FMTL}, \eg, by taking average over clients' models, and apply an $L_2$ regularizer between $\mOmega$ and each $\vw_m$. The corresponding local training step thus could generally be formulated as
\begin{align}
\textbf{Local: } \hspace{4pt} \vw_m^{(t+1)} = \argmin_{\vw} \sL_m(\vw) + \frac{\lambda}{2}\|\vw-\mOmega\|_2^2, \text{ initialized with } \vw_m^{(t)},
\label{eq_pfl_local}
\end{align}
where $\vw_m^{(t)}$ denotes the local model after the $t$-th round; $\lambda$ is the regularization coefficient. \emph{It is worth noting that unlike \autoref{eq_gfl_local}, $\vw$ in \autoref{eq_pfl_local} is initialized by $\vw_m^{(t)}$, not by $\mOmega$ (or $\bar{\vw}$).}

\paragraph{Terminology.} Let us clarify the concepts of ``global'' vs. ``local'' models, and ``generic'' vs. ``personalized'' models. The former corresponds to the \textbf{training} phase: local models are the ones after every round of local training, which are then aggregated into the global model at the server (\autoref{eq_gfl_local}). The latter corresponds to the \textbf{testing} phase:
the generic model is used at the server for generic future test data, while personalized models are specifically used for each client's test data.

\subsection{Local models of generic FL algorithms are strong personalized models}
\label{ss_gfl_good_pfl}

Building upon the aforementioned concepts, we investigate the literature and found that when generic FL algorithms are evaluated in the P-FL setup, it is their global models being tested. In contrast, when personalized FL algorithms are applied, it is their local models (\eg, \autoref{eq_pfl_local}) being tested. This discrepancy motivates us to instead evaluate generic FL algorithms using their local models.

\autoref{fig:overview} summarizes the results (see~\autoref{s_exp} for details). Using local models of \FedAvg (\ie, \autoref{eq_gfl_local}) notably outperforms using its global model in the P-FL setup.
At first glance, this may not be surprising, as local training in \FedAvg is driven by clients' empirical risks. What really surprises us, as will be seen in \autoref{s_exp}, is that \emph{\FedAvg's local models outperform most of the existing personalized FL algorithms, even if no explicit regularization is imposed in \autoref{eq_gfl_local}.}

\subsection{Initialization with weight average is a strong regularizer}
\label{ss_strong_reg}

\begin{wrapfigure}{r}{0.3\columnwidth}
    \centering
    \vspace{-3pt}
    {\includegraphics[width=0.3\textwidth]{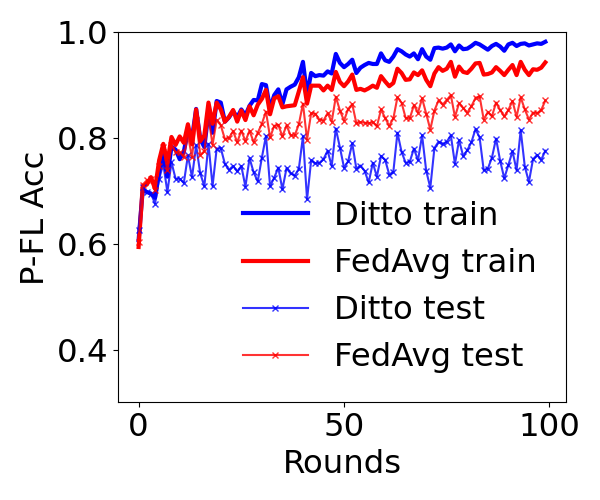}}%
    \vspace{-10pt}
    \caption{\small Comparison of the training and test accuracy in the P-FL setup. \FedAvg's local models achieve lower training accuracy but higher test accuracy.}
    \label{fig:check}    
    \vspace{-30pt}
\end{wrapfigure}

To gain a further understanding, we plot \FedAvg local models' accuracy
on clients' training and test data. We do so also for a state-of-the-art personalized FL algorithm \FMTL~\citep{li2020federated-FMTL}, whose local training step for producing 
personalized models is similar to \autoref{eq_pfl_local}.
As shown in \autoref{fig:check}, \FedAvg has a lower training but higher test accuracy, implying that \FedAvg's local training is more regularized than \autoref{eq_pfl_local}.  

We attribute this effect to the initialization in \autoref{eq_gfl_local}. Specifically, by initializing $\vw$ with $\bar{\vw}$, we essentially impose an $L_2$ regularizer $\frac{\lambda}{2}\|\vw-\bar{\vw}\|_2^2$ with $\lambda\to\infty$ at the beginning of each round of local training, followed by resetting $\lambda$ to be $0$. We found that this implicit regularization leads to a smaller value of $\|\vw-\bar{\vw}\|_2^2$ at the end of each local training round, compared to \autoref{eq_pfl_local}. Due to the page limit, we leave additional analyses in the appendix.
We note that, advanced generic FL algorithms like
\SCAFFOLD~\citep{karimireddy2020scaffold} and \FedDyn~\citep{feddyn} still apply this initialization and learn with the empirical risk during local training. Thus, their local models are strong personalized models as well. 


\section{Federated Robust Decoupling (\FedRoD)}
\label{s_approach}

The fact that personalized models emerge from generic FL algorithms motivate us to focus more on how to improve the latter, especially when clients have non-IID data distributions.

\subsection{Improving generic FL with Balanced Risk Minimization (BRM)}
\label{ss_BSM}
 
We first analyze what factors may lead to non-IID conditions. 
Suppose the data instance $(\vx, y)$ of client $m$ is sampled from a client-specific joint distribution $\sP_m(\vx, y)=\sP_m(\vx|y)\sP_m(y)$, the non-IID distributions among clients can result from non-identical {class distributions} $\sP_m(\vx|y)$, non-identical {class-conditional data distributions} $\sP_m(y)$, or both. All these cases can make $\sL_m(\vw)$ deviate from $\sL(\vw)$ in \autoref{eq:obj}, which is the main cause of degradation in generic FL~\citep{li2020convergence,li2020federated}.

One way to mitigate the influence of non-IID data is to make $\sL_m(\vw)$ align with each other. This can be challenging to achieve if clients have different $\sP_m(\vx|y)$: without knowing clients' data\footnote{Clients having different $\sP_m(\vx|y)$ is related to domain adaptation \citep{gong2012geodesic} and  generalization~\citep{muandet2013domain}, which require knowing the distributions of all/some clients for algorithm design.},
it is hard to design such an aligned $\sL_m(\vw)$.
However, when clients have different $\sP_m(y)$\footnote{This is indeed the main cause of non-IID data distributions in the literature of FL \citep{hsu2019measuring,hsu2020federated}.}, \ie, different and hence imbalanced class distributions, we can indeed design a consistent local training objective by setting a shared goal for the clients --- \emph{the learned local models should classify all the classes well.} It is worth noting that setting such a goal does not require every client to know others' data.

Learning a classifier to perform well on all classes irrespective of the training class distribution is the main focus of class-imbalanced learning~\citep{johnson2019survey,he2009learning,japkowicz2000class}. 
We therefore propose to treat each client's local training as a class-imbalanced learning problem and leverage techniques developed in this sub-field. 
Re-weighting and re-sampling \citep{buda2018systematic} are the most fundamental techniques. Denote by $N_{m,c}$ the number of training instances of class $c$ for client $m$, these techniques adjust $\sL_m(\vw)$ in \autoref{eq:obj} into
\begin{align}
\sL_m^{\textcolor{red}{BR}}(\vw) \propto \sum_{i} {\color{red}q_{y_i}}\ell(\vx_i, y_i; \vw), \hspace{10pt} \text{ where } q_{y_i} \text{ is usually set as } \frac{1}{N_{m,y_i}} \text{ or } \frac{1}{\sqrt{N_{m,y_i}}}.
\label{eq:BL}
\end{align}
Namely, they mitigate the influence of $\sP_m(y)$ by turning the empirical risk $\sL_m$ into a \emph{balanced risk} $\sL_m^{\textcolor{red}{BR}}$, {such that every client solves a more consistent objective that is robust to the class distributions.}
Recently, many class-imbalanced works proposed to replace the instance loss $\ell$ (\eg, cross entropy) with a class-balanced loss \citep{ren2020balanced,cao2019learning,ye2020identifying,khan2017cost,kang2019decoupling}, showing more promising results than re-weighting or re-sampling. We can also define $\sL_m^{\textcolor{red}{BR}}$ using these losses, \eg, the balanced softmax (BSM) loss \citep{ren2020balanced}
\begin{align}
\sL_m^{\textcolor{red}{BR}}(\vw) \propto \sum_{i} \ell^\text{BSM}(\vx_i, y_i; \vw), \hspace{2.5pt}\text{ where }\hspace{2.5pt}
\ell^\text{BSM}(\vx, y; \vw) =  -\log\frac{{\color{red}N^{\gamma}_{m,y}}\exp(g_y(\vx; \vw))}{\sum_{c\in\C} {\color{red}N^{\gamma}_{m,c}} \exp(g_c(\vx; \vw))}.
\label{eq:BSM}
\end{align}
Here, $g_c(\vx; \vw)$ is the logit for class $c$, $\C$ is the label space, and $\gamma$ is a hyper-parameter.
The BSM loss is an unbiased extension of
softmax to accommodate the class distribution shift between training and testing.  
It encourages a minor-class instance to claim a larger logit $g_y(\vx; \vw)$ in training to overcome feature deviation \citep{ye2020identifying} in testing.
We list other class-balanced losses in the appendix. 

We take advantage of these existing efforts by replacing the empirical risk $\sL_m$ in \autoref{eq_gfl_local} with a balanced risk $\sL_m^{\textcolor{red}{BR}}$, which either takes the form of \autoref{eq:BL} or applies a class-balanced loss (\eg, \autoref{eq:BSM}), or both. 
We note that, a variant of \autoref{eq:BL} has been used in \citep{hsu2020federated}. However, our experiments show that it is less effective than class-balanced losses in extreme non-IID cases. 
Interestingly, we found that $\sL_m^{\textcolor{red}{BR}}$ can easily be incorporated into advanced FL algorithms like \FedDyn \citep{feddyn}, because these algorithms are agnostic to the local objectives being used.

\subsection{Local training and local model decoupling with ERM and BRM}
\label{ss_decouple}

The use of balanced risk $\sL_m^{\textcolor{red}{BR}}$ in local training notably improves the resulting global model $\bar{\vw}$'s generic performance,
as will be seen in \autoref{s_exp}. Nevertheless, it inevitably hurts the local model $\vw_m$'s personalized performance, since it is no longer optimized towards client's empirical risk $\sL_m$.

\begin{figure}[t]
    \centering
    \minipage{1\textwidth}
    \centering
    \includegraphics[width=0.5\textwidth]{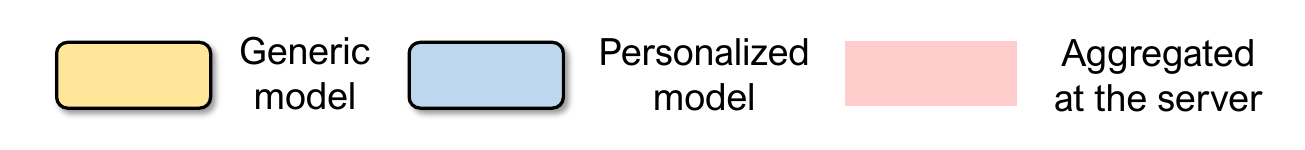}\\
    \endminipage
    \vspace{1pt}
    \minipage{0.49\textwidth}
    \centering
    \includegraphics[width=1\textwidth]{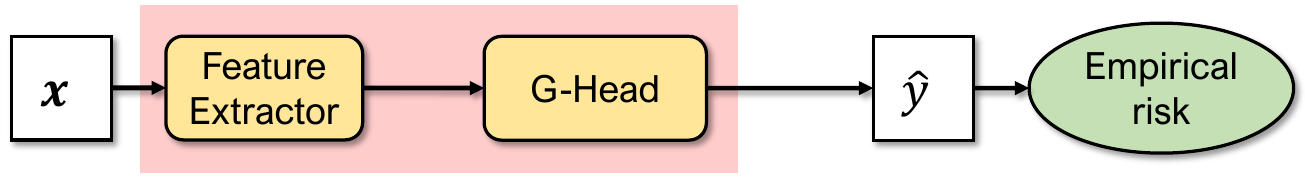}\\
    \mbox{\footnotesize (a) Empirical risk minimization (ERM)}
    \endminipage
    \hfill
    \minipage{0.49\linewidth}
    \centering
      \includegraphics[width=1\textwidth]{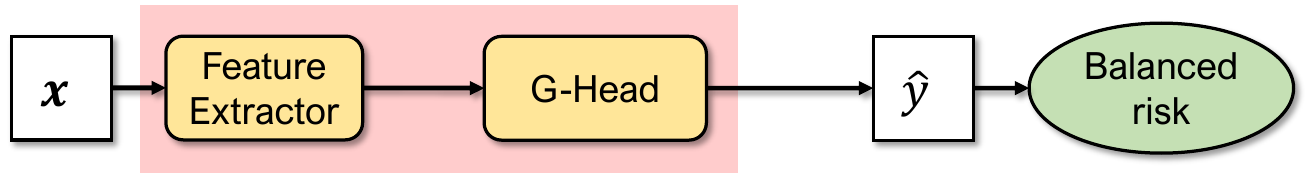}
    \mbox{\footnotesize (b) Balanced risk minimization (BRM)}
    \endminipage
    \vspace{5pt}
    \minipage{0.49\linewidth}
    \centering
     \includegraphics[width=1\textwidth]{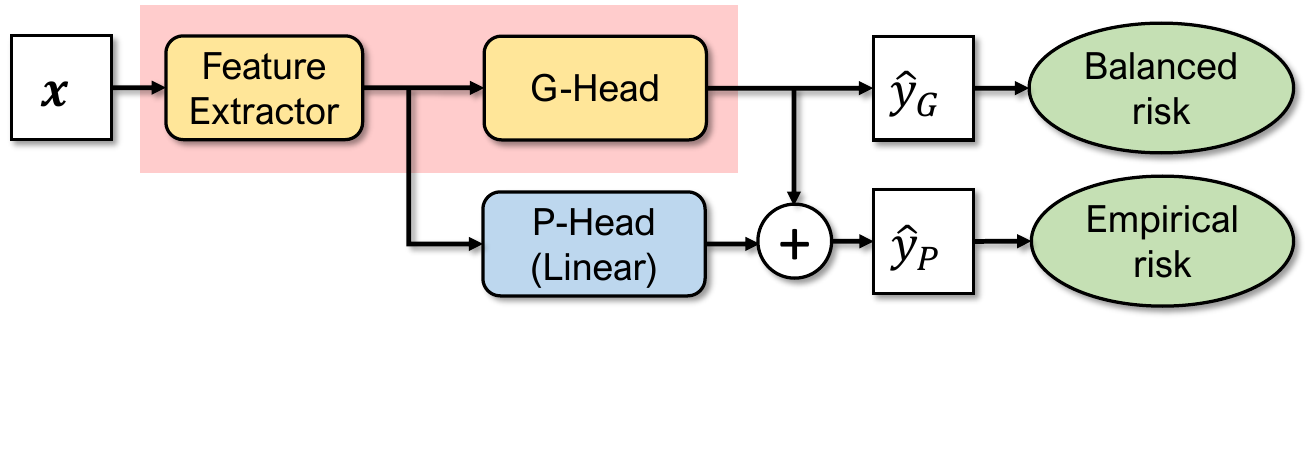}
    \mbox{\footnotesize (c) \FedRoD (linear)}
    \endminipage
    \hfill
    \minipage{0.49\linewidth}
    \centering
     \includegraphics[width=1\textwidth]{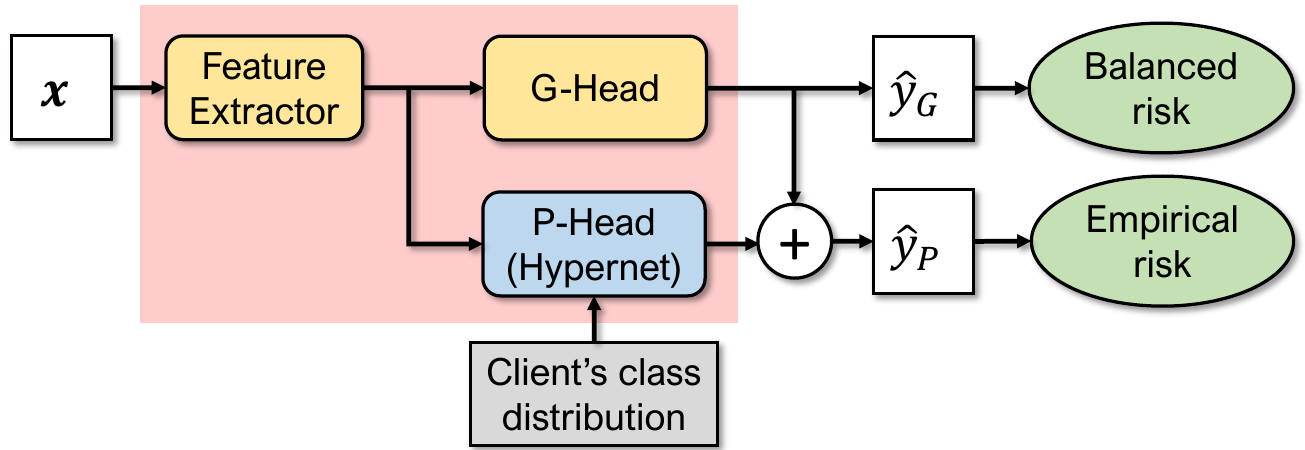}
     \mbox{\footnotesize (d) \FedRoD (hyper)}
    \endminipage
    \vspace{-8pt}
    \caption{\small \textbf{Comparison of local training strategies and model architectures.} G: generic; P: personalized. Yellow boxes correspond to the models for G-FL; green boxes, for P-FL. Boxes that are covered by the red background are sent back to the server for aggregation (\eg, weight average), and re-initialized at the next round. Green ellipsoids correspond to the learning objectives.
    $\hat{y}$ stands for the predicted logits (of all classes); $\hat{y}_G$ and $\hat{y}_P$ come from the G-head and P-head, respectively. \textbf{(a)} local training with ERM; \textbf{(b)} local training with BRM; \textbf{(c)} \textbf{\FedRoD (linear):} learning with both BRM (for G-FL) and ERM (for P-FL) using the two-predictor (head) architecture; \textbf{(d)} \textbf{\FedRoD (hyper):} same as (c), but the P-head is constructed by a shared hypernetwork.} 
    \label{fig:arch}
    \vspace{-11pt}
\end{figure}

To address these contrasting pursuits of generic and personalized FL, we propose a unifying FL framework named \textbf{Federated Robust Decoupling (\FedRoD)}, which \emph{decouples the dual duties of local models} by learning two predictors on top of a shared feature extractor: one trained with empirical risk minimization (ERM) for personalized FL (P-FL) and the other with balanced risk minimization (BRM) for generic FL (G-FL). \autoref{fig:arch} (c-d) illustrates the model and local training objective of \FedRoD.
The overall training process of \FedRoD follows \FedAvg, iterating between local training and global aggregation. As mentioned in \autoref{ss_BSM}, other generic FL algorithms~\citep{feddyn,karimireddy2020scaffold,li2020federated} can easily be applied to the BRM branch to further improve the generic performance. Without loss of generality, we focus on the basic version built upon \FedAvg. We start with  the model in \autoref{fig:arch} (c).

\paragraph{Notations.} We denote by $f(\vx;\vtheta)$ the shared feature extractor parameterized by $\vtheta$, whose output is $\vz$. We denote by $h^G(\vz; \vpsi)$ and $h^P(\vz; \vphi_m)$ the generic and personalized prediction heads parameterized by $\vpsi$ and $\vphi_m$, respectively; both are fully-connected (FC) layers. In short, our generic model is parameterized by $\{\vtheta, \vpsi\}$; our personalized model for client $m$ is parameterized by $\{\vtheta, \vpsi, \vphi_m\}$.

\paragraph{Predictions.} For generic prediction, we perform $\vz=f(\vx;\vtheta)$, followed by $\hat{y}_G = h^G(\vz; \vpsi)$. For personalized prediction, we perform $f(\vx;\vtheta)$, followed by $\hat{y}_P = h^G(\vz; \vpsi)+h^P(\vz; \vphi_m)$. That is, $h^P$ is an add-on to $h^G$, providing personalized information that is not captured by the generic head. 

\paragraph{The overall objective.} \FedRoD learns the generic model with the balanced risk $\sL_m^{\textcolor{red}{BR}}$ and the personalized predictor with the empirical risk $\sL_m$. That is, different from \autoref{eq:obj},
\FedRoD aims to solve the following two optimization problems \emph{simultaneously}
\begin{align}
\min_{\vtheta, \vpsi}~\sL(\{\vtheta, \vpsi\}) = \sum_{m=1}^M \frac{|\sD_m|}{|\sD|} \sL_m^{\text{\color{red}BR}}(\{\vtheta, \vpsi\})
\quad \text{ and }\quad \min_{\vphi_m} \sL_m(\{\vtheta, \vpsi, \vphi_m\}), \forall m \in [M].
\label{eq_fedrod}
\end{align}
We note that, $\sL_m$ is only used to learn the personalized head parameterized by $\vphi_m$.

\paragraph{Learning.} \autoref{eq_fedrod} cannot be solved directly in federated learning, so \FedRoD follows \FedAvg to learn iteratively between the local training and global aggregation steps
\begin{align}
\textbf{Local: } & \hspace{4pt} \vtheta^\star_m, \vpsi^\star_m  = \argmin_{\vtheta, \vpsi} \sL_m^{\text{\color{red}BR}}(\{\vtheta, \vpsi\}),  
&& \text{ initialized with } \bar{\vtheta}, \bar{\vpsi} \label{eq_ours_local_g}, \\
& \hspace{4pt} \vphi^\star_m  = \argmin_{\vphi_m} \sL_m(\{\vtheta, \vpsi,\vphi_m\}), 
&& \text{ initialized with } \vphi'_m,
\label{eq_ours_local_p} \\
\textbf{Global: } & \hspace{4pt}\bar{\vtheta} \leftarrow \sum_{m=1}^M \frac{|\sD_m|}{|\sD|}{\vtheta_m^\star}, \hspace{5pt} \bar{\vpsi} \leftarrow \sum_{m=1}^M \frac{|\sD_m|}{|\sD|}{\vpsi_m^\star}, &&
\label{eq__ours_global}
\end{align}
where $\vphi'_m$ is learned from the previous round, similar to $\vw_m^{(t)}$ in \autoref{eq_pfl_local}. 
That is, the personalized head will not be averaged globally but kept locally. In our implementation, \autoref{eq_ours_local_g} and \autoref{eq_ours_local_p} are solved simultaneously via SGD, and we do not derive gradients w.r.t. $\vtheta$ and $\vpsi$ from $\sL_m(\{\vtheta, \vpsi, \vphi_m\})$.
The $\vtheta$ and $\vpsi$ in \autoref{eq_ours_local_p} thus come dynamically from the SGD updates of \autoref{eq_ours_local_g}. In other words, \autoref{eq_ours_local_p} is
not merely fine-tuning on top of the generic model. 
In the end of federated learning, {we will obtain $\bar{\vtheta}$ and $\bar{\vpsi}$ (\autoref{eq__ours_global}) for generic predictions and $\{\vtheta^\star_m, \vpsi^\star_m, \vphi^\star_m\}_{m=1}^M$ (\autoref{eq_ours_local_g} and \autoref{eq_ours_local_p}) for personalized predictions, respectively.} Please be referred to the appendix for the pseudocode.

\subsection{Adaptive personalized predictors via hypernetworks} \label{ss_hype}

In \autoref{ss_decouple}, the parameter $\vphi_m$ of the personalized predictor is learned independently for each client and never shared across clients. In other words, for a new client not involved in the training phase, \FedRoD can only offer the global model for generic prediction.
In this subsection, we investigate learning a shared personalized predictor that can adapt to new clients. 
Concretely, we propose to learn a \emph{meta-model} which can generate $\vphi_m$ for a client given the client's class distribution. We denote by $H^P(\vct{a}_m; \vnu)$ the meta-model parameterized by $\vnu$, whose output is $\vphi_m$. Here, $\vct{a}_m \in \R^{|\C|}$ is the $|\C|$-dimensional vector that records the class distribution of client $m$; \ie, the $c$-th dimension $a_m[c]=\frac{N_{m,c}}{\sum_{c'} N_{m,c'}}$.
Accordingly, the local training step of $\vphi_m$ in \autoref{eq_ours_local_p} is replaced by
\begin{align}
\textbf{Local: } \hspace{4pt} \vnu^\star_m  = \argmin_{\vnu} \sL_m(\{\vtheta, \vpsi, \vnu\}), \text{ initialized with } \bar{\vnu}; \hspace{8pt} \textbf{Global: }  \hspace{4pt} \bar{\vnu} \leftarrow \sum_{m=1}^M \frac{|\sD_m|}{|\sD|}{\vnu_m^\star}. \label{eq_meta_hyper}
\end{align}
We implement $H^P$ by a lightweight hypernetwork \citep{ha2016hypernetworks} with two fully-connected layers. With the learned $\bar{\vnu}$, the meta-model $H^P$ can locally generate $\vphi_{m}$ based on $\vct{a}_{m}$, making it adaptive to new clients simply by class distributions. The parameter $\vphi_{m}$ can be further updated using clients' data. \textbf{We name this version \FedRoD (hyper); the previous one, \FedRoD (linear).} Please see \autoref{fig:arch} (c-d) for an illustration. We include more details in the appendix. 


\section{Experiment (More Details and Results in The Appendix)}
\label{s_exp}

\definecolor{Gray}{gray}{0.9}
\begin{table*}[t!] 
    \scriptsize
	\centering
	\caption{\small Results in G-FL accuracy and P-FL accuracy ($\%$). $\star$: methods with no G-FL models and we combine their P-FL models. $\mathsection$: official implementation. \textcolor{blue}{Blue}/\textbf{bold} fonts highlight the best baseline/our approach.} 
	\vspace{-5pt}
	\setlength{\tabcolsep}{1.25pt}
	\renewcommand{\arraystretch}{0.35}
	\begin{tabular}{l|ccc|ccc|ccc|ccc|ccc|ccc|ccc|ccc|ccc|ccc|ccc}
	\toprule {Dataset} &
	\multicolumn{3}{c|}{EMNIST}&
	\multicolumn{6}{c|}{FMNIST}& \multicolumn{6}{c|}{CIFAR-10} & \multicolumn{6}{c|}{CIFAR-100}\\
	\midrule {Non-IID} &
	\multicolumn{3}{c|}{Writers
} & \multicolumn{3}{c|}{Dir(0.1)
} & \multicolumn{3}{c}{Dir(0.3)} & \multicolumn{3}{|c|}{Dir(0.1)
} & \multicolumn{3}{c}{Dir(0.3)} & \multicolumn{3}{|c|}{Dir(0.1)
} & \multicolumn{3}{c|}{Dir(0.3)}\\
    \midrule
    {Test Set} & G-FL & \multicolumn{2}{|c|}{P-FL} & G-FL & \multicolumn{2}{|c|}{P-FL} & G-FL & \multicolumn{2}{|c|}{P-FL} & G-FL & \multicolumn{2}{|c|}{P-FL} & G-FL & \multicolumn{2}{|c|}{P-FL} & G-FL & \multicolumn{2}{|c|}{P-FL} & G-FL & \multicolumn{2}{|c|}{P-FL}\\
    \midrule
    
    {Method / Model} &
    \multicolumn{1}{c|}{GM} & GM & PM & \multicolumn{1}{c|}{GM} & GM & PM & \multicolumn{1}{c|}{GM} & GM & PM & \multicolumn{1}{c|}{GM} & GM & PM & \multicolumn{1}{c|}{GM} & GM & PM & \multicolumn{1}{c|}{GM} & GM & PM & \multicolumn{1}{c|}{GM} & GM & PM\\
    \midrule

    \FedAvg& 97.0 & 96.9 & 97.2 & 81.1 & 81.0 & \textcolor{blue}{91.5} & 83.4 &83.2 & 90.5 & 57.6 & 57.1 & 90.5 & 68.6 & 69.4 & 85.1 & 41.8 &	41.6 &	70.2 &	46.4&	46.2&	61.7\\ 

    \FedProx& 97.0 & 97.0 & 97.0 & 82.2 & 82.3& 91.4 & 84.5 & 84.5 & 89.7 & 58.7 & 58.9 & 89.7 & 69.9 & 69.8 & 84.7 & 41.7 & 41.6 & 70.4 & 46.5 & 46.4 & 61.5\\

    \SCAFFOLD& 97.1 & 97.0 & 97.1 & 83.1 & 83.0 & 89.0 & 85.1 & 85.0 & 90.4 & 61.2 & 60.8 & 90.1 & 71.1 & 71.5 & 84.8 & 42.3 & 42.1 & 70.4 & 46.5 & 46.5 & 61.7\\

    \FedDyn$\mathsection$& \textcolor{blue}{97.3}& \textcolor{blue}{97.3} &97.3 & \textcolor{blue}{83.2} & \textcolor{blue}{83.2} &90.7 & \textcolor{blue}{86.1} & \textcolor{blue}{86.1} &  \textcolor{blue}{91.5} & \textcolor{blue}{63.4} & \textcolor{blue}{63.9} & \textcolor{blue}{92.4} & \textcolor{blue}{72.5} & \textcolor{blue}{73.2} & \textcolor{blue}{85.4} & \textcolor{blue}{43.0} & \textcolor{blue}{43.0} & \textcolor{blue}{72.0} & \textcolor{blue}{47.5} & \textcolor{blue}{47.4} & \textcolor{blue}{62.5}\\
    \midrule

    \MOCHA$^\star$ & 75.4& 75.0& 85.6& 36.1 & 36.0 & 87.3 & 53.1 & 53.4 & 78.3 & 12.1 & 12.7 & 90.6 & 13.5 & 13.7 & 80.2 & 9.5 & 9.3 & 60.7 & 10.8 & 10.7 & 49.9\\

    \LGFedAvg$^\star\mathsection$ & 80.1 & 80.0 & 95.6 &54.8 & 54.5 & 89.5 & 66.8 & 66.8 & 84.4 & 29.5 & 28.8 & 90.8 & 46.7 & 46.2 & 82.4 & 23.5 & 23.4 & 66.7 & 34.5 & 33.9 & 55.4 \\

    \FedPer$^\star$& 93.3 & 93.1 & 97.2 & 74.5&74.4&91.3 & 79.9&79.9&90.4& 50.4&50.2&89.9 & 64.4&64.5&84.9 & 37.6 &	37.6&	71.0&	40.3&	40.1&	\textcolor{blue}{62.5}\\

    \PerFedAvg & 95.1 & - & 97.0 & 80.5 & - & 82.8 & 84.1 & - & 86.7 & 60.7 & - & 82.7 & 70.5& - & 80.7 & 39.0 & - & 66.6 & 44.5 & - & 58.9\\

    \pFedMe$\mathsection$ & 96.3 & 96.0 & 97.1 & 76.7 & 76.7 & 83.4 & 79.0 & 79.0 & 83.4 & 50.6 & 50.7 & 76.6 & 62.1 & 61.7 & 70.5 & 38.6 & 38.5 & 63.0 & 41.4 & 41.1 & 53.4\\

    \FMTL & 97.0& 97.0& 97.4& 81.5 & 81.5 & 89.4 & 83.3 & 83.2 & 90.1 & 58.1 & 58.3 & 86.8 & 69.7 & 69.8 & 81.5 & 41.7 & 41.8 & 68.5 & 46.4 & 46.4 & 58.8\\ 
    \FedFOMO$^\star$ & 80.5 & 80.4 & 95.9 & 34.5 & 34.3 & 90.0 & 70.1 & 69.9 & 89.6 & 30.5 & 31.2 & 90.5 & 45.3 & 45.1 & 83.4 & 35.4 & 35.3 & 68.9 & 39.6 & 39.3 & 58.4\\

    \FedRep$^\star\mathsection$ & 95.0 & 95.1 & \textcolor{blue}{97.5} & 79.5 & 80.1 & 91.8 & 80.6 & 80.5 & 90.5 & 56.6 & 56.2 & 91.0 & 67.7 & 67.5 & 85.2 & 40.7 & 40.7 & 71.5 & 46.0 & 46.0 & 62.1\\
    \midrule
    Local only & - & - & 64.2 & - & - & 85.9 & - & - & 85.0 & - & - & 87.4 & - & - & 75.7 & - & - & 40.0 & - & - & 32.5\\
    \midrule
    \textbf{\FedRoD (linear)} & \textbf{97.3} & \textbf{97.3} & \textbf{97.5} & \textbf{83.9} & \textbf{83.9} & \textbf{92.7} & \textbf{86.3} & \textbf{86.3} & \textbf{94.5} & \textbf{68.5} & \textbf{68.5} & \textbf{92.7} & \textbf{76.9} & \textbf{76.8} & \textbf{86.4} & \textbf{45.9} &\textbf{45.8} &	\textbf{72.2} &	\textbf{48.5} &	\textbf{48.5} &	\textbf{62.3}\\
    \textbf{\FedRoD (hyper)} & \textbf{97.3} & \textbf{97.3} & \textbf{97.5} & \textbf{83.9} & \textbf{83.9} & \textbf{92.9} & \textbf{86.3} & \textbf{86.3} & \textbf{94.8} & \textbf{68.5} & \textbf{68.5} & \textbf{92.5} & \textbf{76.9} & \textbf{76.8} & \textbf{86.8} & \textbf{45.9} &	\textbf{45.8} &	\textbf{72.3} &	\textbf{48.5} &	\textbf{48.5} &	\textbf{62.5} 
\\
    \quad\quad + \FedDyn & \textbf{97.4} & \textbf{97.4} & \textbf{97.5} & \textbf{85.9} & \textbf{85.7} & \textbf{95.3} & \textbf{87.5} & \textbf{87.5} & \textbf{94.6} & \textbf{68.2} & \textbf{68.2} & \textbf{92.7} & \textbf{74.6} & \textbf{74.6} & \textbf{85.6} & 
    \textbf{46.2} &	\textbf{46.2} &	\textbf{72.5} &	\textbf{48.4} &	\textbf{48.4} &	\textbf{62.5}\\
    \bottomrule
	\end{tabular}
	\label{tbl:main_bal}
	\vspace{-14pt} 
\end{table*}

\noindent\textbf{Datasets, models, and settings.} We use CIFAR-10/100~\citep{krizhevsky2009learning} and Fashion-MNIST (FMNIST)~\citep{Xiao2017FashionMNISTAN}. We also include a realistic EMNIST~\citep{cohen2017emnist} dataset, which collects hand-written letters of thousands of writers. 
To simulate the non-IID data distributions on CIFAR and FMNIST, we follow \cite{hsu2019measuring} to create a heterogeneous partition for $M$ clients: an $M$-dimensional vector $\vq_c$ is drawn from $\text{Dir}(\alpha)$ for class $c$, and we assign data of class $c$ to client $m$ proportionally to $\vq_c[m]$. The resulting clients have different numbers of total images and different class distributions. {With $\alpha < 1$, most of the training examples of one class are likely assigned to a small portion of clients. Similar to~\cite{lin2020ensemble}, we use $M=100$ clients for FMNIST and $M=20$ for CIFAR-10/100, and sample $20\%/40\%$ clients at every round, respectively. For EMNIST, we use the digit images, follow~\cite{caldas2018leaf} to construct $2,185$ clients (each is a writer), and sample $5\%$ clients at every round.} We use a ConvNet~\citep{lecun1998gradient} similar to \citep{mcmahan2017communication,feddyn}. It contains $3$ convolutional layers and $2$ fully-connected layers. We train every FL algorithm for $100$ rounds, with $5$ local epochs in each round.
 
We report the mean accuracy of \textbf{five times} of experiments with different random seeds. We evaluate the generic performance (G-FL) using the \textbf{generic model (GM)} on the standard generic test set. For FMNIST and CIFAR-10/100, we evaluate the personalized performance (P-FL) using \textbf{personalized models (PM)} on the same set, but re-weight the accuracy according to clients' class distributions $P_m(y)$ and average the weighted accuracy across $M$ clients {as $\frac{1}{M}\sum_m\frac{\sum_i\sP_m(y_i)\textbf{1}(y_i = \hat y_{i})}{\sum_i\sP_m(y_i)}$}. Here, $i$ is the instance index. This evaluation is more robust (essentially as the expectation) than assigning each client a specific test set. For EMNIST, each client has its own test set with the same writing style.

\noindent\textbf{Our variants.} We mainly use \autoref{eq:BSM} with $\gamma = 1$ as the $\sL_m^{{BR}}$ and report the \FedRoD (hyper) version (cf. \autoref{ss_hype}). \autoref{tbl:ablation_irm} provides the ablation study.

\noindent\textbf{Baselines.}
For G-FL methods including \FedAvg \citep{mcmahan2017communication}, \FedProx \citep{li2020federated}, \SCAFFOLD \citep{karimireddy2020scaffold}, and \FedDyn~\citep{feddyn}, we use their global models $\bar{\vw}$ for G-FL evaluation; their local models (\ie, $\vw_m$ in \autoref{fig:overview}) for P-FL evaluation.
For P-FL methods, to evaluate their G-FL performance, we use the available global models in \pFedMe~\citep{dinh2020personalized} and \FMTL~\citep{li2020federated-FMTL} or average the final personalized models for \MOCHA~\citep{smith2017federated}, \FedPer~\citep{arivazhagan2019federated}, \LGFedAvg~\citep{liang2020think}, \FedFOMO~\citep{fedfomo}, and \FedRep~\citep{collins2021exploiting}.

\emph{To illustrate the difference between applying GMs and PMs in a P-FL setting, we also evaluate the P-FL performance using GMs, which is how \FedAvg has been applied to P-FL in literature.} 
\subsection{Results}

\label{ss_exp_main}

\noindent\textbf{\FedRoD bridges G-FL and P-FL and consistently outperforms all generic and personalized FL methods.} \autoref{tbl:main_bal} summarizes the results. In terms of \textbf{G-FL accuracy}, advanced local training (\ie, \SCAFFOLD, \FedProx, and \FedDyn) outperforms \FedAvg and personalized methods, and our \FedRoD can have further gains by using balanced risk minimization (BRM). We also investigate combining \FedRoD and \FedDyn~\citep{feddyn}, using the latter to optimize the generic model with BRM, which outperforms either ingredient in many cases.
We report the G-FL accuracy of personalized FL algorithms mainly to investigate if they have similar properties like \FedAvg: an algorithm designed for one setup can also construct models for the other setup. 

In terms of \textbf{P-FL accuracy}, by using PMs most methods outperform the baseline of local training with individual client's data without communication (\ie, local only), justifying the benefits of federated collaboration\footnote{PMs of ``local only'' could outperform the GM of \FedAvg on the P-FL accuracy, especially when the non-IID condition becomes severe (\eg, Dir (0.1)): it is hard to train a single GM to perform well in P-FL.}. 
For generic FL methods, using PMs (\ie, local models $\{\vw_m\}$) clearly outperforms using GMs (\ie, $\bar{\vw}$), which supports our claims and observations in \autoref{fig:overview} and \autoref{ss_gfl_good_pfl}.
\emph{It is worth noting that the local models from generic FL methods are highly competitive to or even outperform personalized models produced by personalized FL methods.}
This provides generic FL methods with an add-on functionality to output personalized models by keeping the checkpoints on clients after local training.
\textbf{Our \FedRoD achieves the highest P-FL accuracy} and we attribute this to (a) the shared feature extractor learned with the balanced risk and re-initialized every round to benefit from implicit regularization; (b) the personalized head learned with clients’ empirical risks.

 \begin{wrapfigure}{r}{0.36\columnwidth}
    \centering
    \minipage{0.4\columnwidth}   
    \centering
    \vskip -5pt
    \includegraphics[width=0.9\linewidth]{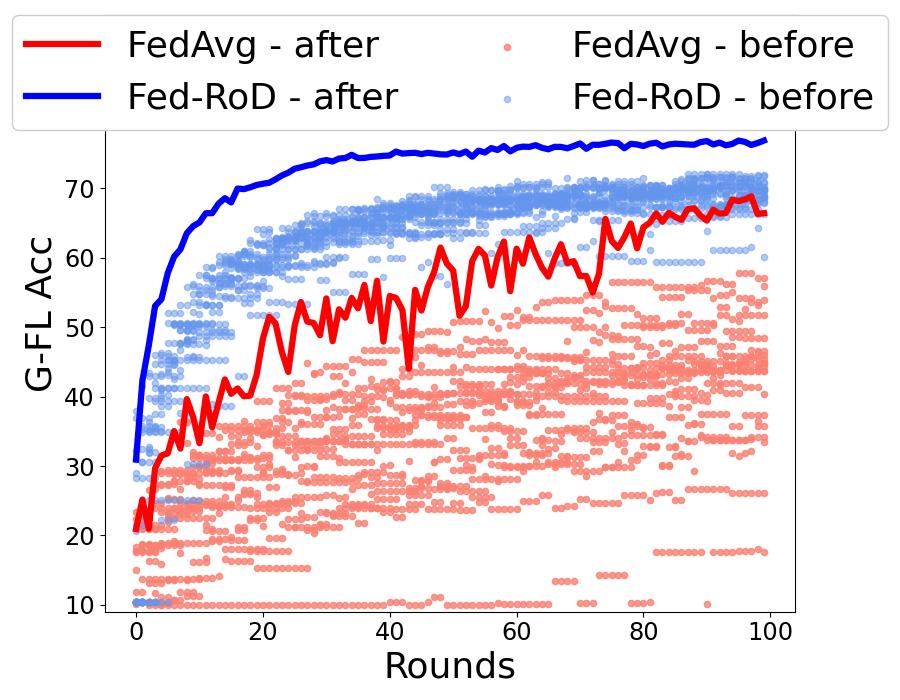}
    \vskip -15pt
    \endminipage\hfill
    \minipage{0.35\columnwidth}
    \centering
    \includegraphics[width=1.0\linewidth]{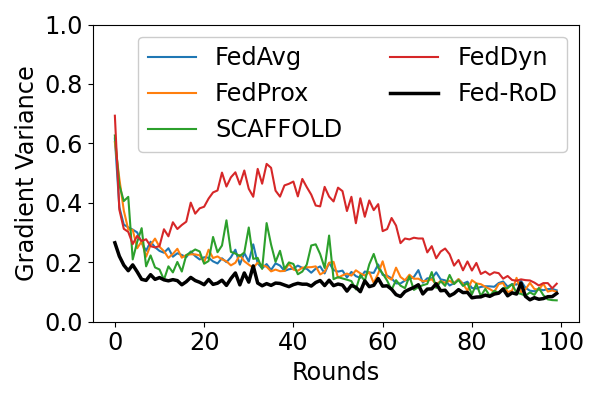}
    \vskip -35pt
    \endminipage\hfill
    \vskip -10pt
    \caption{\small \textbf{Upper:} G-FL test accuracy along the training rounds before/after averaging the local models. \textbf{Lower:} variances of $\vw_m-\bar{\vw}$ across clients.}
    \label{fig:analysis}
    \vskip -35pt
\end{wrapfigure}

\noindent\textbf{BRM effectively reduces the variance of G-FL accuracy and local gradients.} To understand why \FedRoD improves G-FL, we visualize the global model $\bar\vw$'s and each local model $\vw_m$'s G-FL accuracy on CIFAR-10 (Dir(0.3)), in~\autoref{fig:analysis} (upper). \FedRoD not only learns a better global model for G-FL, but also has a smaller variance of accuracy across the local models' generic heads (as their objectives are more aligned). We also show how $\vw_m$ deviates from $\bar{\vw}$ after local training in \autoref{fig:analysis} (lower). \FedRoD has a smaller variance. {This coincides with the study in~\citep{kong2021consensus}: lower variances of the local gradients could imply better generic performance.} 

\paragraph{\FedRoD benefits from decoupling.} We compare several variants of \FedRoD (cf.~\autoref{fig:arch}), with one head (reduced to \FedAvg) or different networks (linear/hyper). We evaluate on CIFAR-10 (Dir(0.3)). As shown in~\autoref{tbl:ablation_irm}, \emph{\FedAvg with BRM significantly improves G-FL but degrades in P-FL.} \FedRoD remedies this by training a decoupled personalized head. We note that, \FedRoD does not merely fine-tune the global model with clients' data (cf. \autoref{ss_decouple}). We also compare different balanced losses in~\autoref{tbl:ablation_loss}: advanced losses outperforms  importance re-weighting~\citep{hsu2020federated}.

 \begin{wrapfigure}{r}{0.36\columnwidth}
    \centering
    \vskip -5pt
    \includegraphics[width=0.95\linewidth]{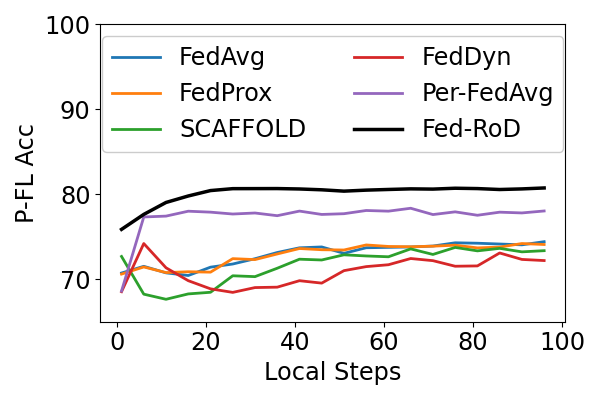}
    \vskip -10pt
    \caption{\small The average P-FL accuracy on future clients, with local training.}
    \label{fig:new_clients}
    \vskip -10pt
\end{wrapfigure}

\paragraph{\FedRoD (hyper) benefits future clients.} To validate the generalizability to new clients, we build on the Dir(0.3) non-IID setting for FMNIST and CIFAR-10/100, but split the training data into $100$ clients ($50$ are in training; $50$ are new). We train on the $50$ training clients for $100$ rounds (sampling $20$ of them every round). We then evaluate on the 50 new clients individually, either using the global model directly or fine-tuning it with clients' data for several steps.~\autoref{tbl:all_local_bf_aft} and \autoref{fig:new_clients} shows the averaged accuracy on new clients. Without fine-tuning, \FedRoD (hyper) can already generate personalized models, and outperforms others methods stably with fine-tuning.

\begin{table}[] 
	\centering
	\begin{minipage}{0.3\linewidth}\centering
	\scriptsize
	\caption{\small Ablation study on variants of \FedRoD. FT: fine-tuning}
	\vspace{-8pt}
	\setlength{\tabcolsep}{1pt}
	\renewcommand{\arraystretch}{0.35}
	\begin{tabular}{l|ccc}
    \toprule
    {Test Set} & G-FL & \multicolumn{2}{|c}{P-FL} \\
    \midrule
    {Method / Model} & \multicolumn{1}{c|}{GM} & GM & PM \\
    \midrule
    Centralized &  85.4 & 85.4 & -\\
    \midrule
    \FedAvg & 68.6 & 69.4 & 85.1\\
    \FedAvg (BRM) & 76.8 & 76.7 & 76.1 \\ 
    \FedAvg (BRM, FT) & 76.8 & 76.7 & 84.5 \\
    \FedRoD (linear, BRM) & 76.9 & 76.8 & 86.4\\
    \FedRoD (hyper, BRM) & 76.9 & 76.8 &86.8 \\
    \bottomrule
	\end{tabular}
	\label{tbl:ablation_irm}
	\end{minipage}
    \hfill
    \begin{minipage}{0.28\linewidth}\centering
	\scriptsize
	\caption{\small \FedRoD with different balanced losses. $\star$: BSM} 
	\vspace{1pt}
	\setlength{\tabcolsep}{1pt}
	\renewcommand{\arraystretch}{0.35}
	\begin{tabular}{l|ccc}
    \toprule
    {Test Set} & G-FL & \multicolumn{2}{|c}{P-FL} \\
    \midrule
    {Loss / Model} & \multicolumn{1}{c|}{GM} & GM & PM \\
    \midrule
    Cross entropy & 68.6 & 69.4 & 85.1\\
    \citep{hsu2020federated} & 65.8 & 65.8 & 80.1 \\
    \citep{cao2019learning} & 75.7 & 75.9 & 83.3\\
    \citep{ye2020identifying} & 75.2 & 75.0 & 85.1 \\
    \citep{ren2020balanced}$\star$ &  76.9 & 76.8 & 86.8\\
    \bottomrule
	\end{tabular}
	\label{tbl:ablation_loss}
    \end{minipage}
    \hfill
    \begin{minipage}{0.375\linewidth}\centering
	\scriptsize
	\caption{\small P-FL accuracy on future non-IID clients (Dir(0.3) for all datasets). Each cell is {before}/{after} local fine-tuning.} 
	\vspace{-7pt}
	\setlength{\tabcolsep}{0.87pt}
	\renewcommand{\arraystretch}{0.35}
	\begin{tabular}{l|ccc}
	\toprule
	Method &  FMNIST &  CIFAR-10 & CIFAR-100\\
    \midrule
    \FedAvg & 80.3/87.2 & 56.2/76.0 & 38.9/54.3 \\
    \FedDyn & 82.3/87.6 & 61.7/76.2 & 40.1/57.2\\
    \PerFedAvg& 82.1/89.6 & 60.0/79.8 & 37.6/55.6\\
    \midrule
    \FedRoD (linear) & 83.5/91.3 & 62.4/80.2 & 40.0/58.2 \\
    \FedRoD (hyper) & 88.9/91.4 & 75.7/81.5 & 40.7/59.0\\
    \quad\quad+\FedDyn & 89.2/91.3 & 77.1/83.5 & 41.4/59.5\\
    \bottomrule
	\end{tabular}
	\label{tbl:all_local_bf_aft}
    \end{minipage}
    \vspace{-12pt}
\end{table}

\paragraph{More results and analyses in the Appendix.} The Appendix includes studies with class-imbalanced global distributions and we show that \FedRoD still performs well. BRM can be further improved with meta-learned hyper-parameters. We validate that re-initializing the local models by the global model at every round (\ie, \autoref{eq_gfl_local}) does lead to a much smaller regularization loss than \autoref{eq_pfl_local} to support our claim in~\autoref{ss_strong_reg}. More comprehensive results regarding more clients, deeper backbones, compatibility with other methods, robustness against adversaries, etc, are also provided.


\section{Conclusion}
\label{s_disc}

Most of the existing work in federated learning (FL) has been dedicated to either learning a better generic model or personalized models. We show that these two contrasting goals can be achieved simultaneously via a novel two-loss, two-predictor FL framework \FedRoD. Concretely, we show that strong personalized models emerge from the local training of generic FL algorithms, due to implicit regularization; imposing class-balanced objectives further improves the generic FL accuracy when clients have non-IID distributions. \FedRoD seamlessly incorporates these two observations to excel in both FL settings, and further enables fast adaptation to new clients via an adaptive module.

\section*{Acknowledgments}
This research is partially supported by NSF IIS-2107077, NSF OAC-2118240, NSF OAC-2112606, and the OSU GI Development funds. We are thankful for the generous support of the computational resources by the Ohio Supercomputer Center and AWS Cloud Credits for Research.

\section*{Reproducibility Statement}
We report the results with the average over $5$ runs of different random seeds. We exhaustively provide the information about the hyperparameters, datasets, evaluation, and other details in~\autoref{s_exp} and~\autoref{suppl-sec:exp_s}, which should be comprehensive for reproducibility. We also provide our code in~\url{https://github.com/hongyouc/Fed-RoD}.

{\small
\bibliography{iclr2022_conference}
\bibliographystyle{iclr2022_conference}
}


\clearpage
\appendix
\begin{center}
\textbf{\Large Appendix}
\end{center}

We provide details omitted in the main paper. 
\begin{itemize}
    \item \autoref{suppl-sec:related}: additional comparison to related work (cf. \autoref{s_related} and \autoref{s_GFLPFL} of the main paper).
    \item \autoref{suppl-sec:full}: additional details of \FedRoD (cf. \autoref{s_GFLPFL} and \autoref{s_approach} of the main paper).
    \item \autoref{suppl-sec:exp_s}: details of experimental setups (cf. \autoref{s_exp} of the main paper).
    \item \autoref{suppl-sec:exp_r}: additional experimental results and analysis (cf. \autoref{s_GFLPFL} and \autoref{s_exp} of the main paper). 
\end{itemize}

\section{Comparison to Related Work}
\label{suppl-sec:related}

\subsection{Follow-up works of \FedAvg} 
Several recent works \citep{karimireddy2020scaffold,karimireddy2020mime,zhao2018federated} have shown that, with multiple steps of local SGD updates, the local model $\vw_m$ would drift away from each other, leading to a degenerated global model $\bar{\vw}$ that deviates from the solution of \autoref{eq:obj} of the main text.

One way to mitigate this is to modify the local training objective (cf. \autoref{eq_gfl_local} of the main text). For instance, \FedProx \citep{li2020federated} introduced a regularizer with respect to $\bar{\vw}$,
\begin{align}
\min_{\vw} \sL_m(\vw) + \frac{\lambda}{2}\|\vw-\bar{\vw}\|^2. \nonumber
\end{align}
\FedDyn \citep{feddyn} further added a dynamic term based on the local model of the previous round $\vw'_m$,
\begin{align}
\min_{\vw} \sL_m(\vw) + \langle\nabla\sL_m(\vw'_m),\vw\rangle +  \frac{\lambda}{2}\|\vw-\bar{\vw}\|^2. \nonumber
\end{align}
These regularizers aim to stabilize local training and align the objectives among clients. Some other works did not change the objectives but introduced control variates or momentum to correct the local gradient~\citep{karimireddy2020mime,karimireddy2020scaffold}, or designed a new optimizer more suitable for decentralized learning~\citep{malinovskiy2020local,yuan2020federated,pathak2020fedsplit}.

It is worth mentioning, in most of these works, the empirical risk $\sL_m(\vw)$ still plays an important role in driving the local model update. Since $\sL_m(\vw)$ directly reflects the (non-IID) client data distribution, the learned local models are indeed strong candidates for personalized models.

\subsection{Generic federated learning}
\FedAvg~\citep{mcmahan2017communication} is the standard algorithm, which involves multiple rounds of local training and global aggregation. 
Many works have studied its convergence \citep{khaled2020tighter,haddadpour2019convergence}, robustness \citep{bonawitz2019towards}, communication \citep{konevcny2016federated,reisizadeh2019fedpaq}, especially for non-IID clients \citep{li2020convergence,zhao2018federated,li2020federated}. Many other works proposed to improve \FedAvg. In terms of \textbf{global aggregation}, \citep{yurochkin2019bayesian,wang2020federated} matched local model weights before averaging. \citep{lin2020ensemble,he2020group,zhou2020distilled,chen2021fedbe} replaced weight average by model ensemble and distillation. \citep{hsu2019measuring,reddi2021adaptive} applied server momentum and adaptive optimization to improve the global model update.
In terms of \textbf{local training}, \citep{malinovskiy2020local,yuan2020federated,pathak2020fedsplit,liang2019variance} improved the optimizer. To reduce local models' drifts from the global model, 
\citep{zhao2018federated} mixed client and server data in local training; \FedProx \citep{li2020federated}, \FedDANE \citep{li2019feddane}, and \FedDyn \citep{feddyn} employed regularization toward the global model; \SCAFFOLD~\citep{karimireddy2020mime} \Mime \citep{karimireddy2020scaffold} leveraged control varieties and/or server statistics to correct local gradients; \citep{wang2020tackling,yao2019federated} modified the local model update rules. For most of them, the empirical risks on clients' data are the major forces to drive local training.

We also aim to reduce local models' drifts but via a different way. We directly bypass the empirical risks that reflect clients' data distributions. Instead, we apply objective functions in class-imbalanced learning \citep{he2009learning}, which are designed to be robust to the change of class distributions. Our approach is different from \citep{wang2020towards,yang2020federated,duan2020self}, which monitored and resolved class imbalance from the server while we tackled it at the clients.
Our approach is also different from agnostic FL \citep{mohri2019agnostic,deng2020distributionally}, whose local training is still built on empirical risk minimization. 
The closest to ours is \citep{hsu2020federated}, which used a traditional class-imbalanced treatment, re-weighting, to mitigate non-identical class distributions. We show that more advanced techniques can be applied to further improve the performance, especially under extreme non-IID conditions where re-weighting is less effective.
Moreover, our method is compatible with existing efforts like \FedDyn \citep{feddyn} and \SCAFFOLD~\citep{karimireddy2020mime} to boost the generic performance. 

\subsection{Personalized federated learning}

Personalized FL \citep{Kulkarni2020SurveyOP} learns a customized model for each client. Many approaches are based on multi-task learning (MTL)~\citep{zhang2017survey,ruder2017overview,evgeniou2004regularized,evgeniou2007multi,jacob2009clustered,zhang2010convex} --- leveraging the clients' task relatedness to improve model generalizability. For instance, \citep{smith2017federated} encouraged related clients to learn similar models; \citep{li2020federated-FMTL,dinh2020personalized,hanzely2020lower,hanzely2020federated,corinzia2019variational,li2019fedmd} regularized local models with a learnable global model, prior, or set of data logits. \citep{liang2020think,li2021fedbn,bui2019federated,arivazhagan2019federated} designed the model architecture to have both personalized (usually the feature extractor) and shareable components. \citep{fedfomo,Huang2021PersonalizedCF} constructed for each client an initialized model or regularizer based on learnable bases. Our approach is inspired by MTL as well but has several notable differences from existing works. First, we found that the global aggregation step in generic FL already serves as a strong regularizer.
Second,
instead of learning for each client a personalized feature extractor \citep{bui2019federated,liang2020think} or an entire independent model that can operate alone \citep{smith2017federated,dinh2020personalized,hanzely2020lower}, \FedRoD shares a single feature extractor among all clients, inspired by invariant risk minimization \citep{arjovsky2019invariant,ahuja2020invariant} and domain generalization \citep{muandet2013domain,ghifary2015domain}. This reduces the total parameters to be learned and improves model's generalizability. Compared to \FedPer~\citep{arivazhagan2019federated} and \FedRep~\citep{collins2021exploiting} which also learned a shared feature extractor, \FedRoD simultaneously outputs a single, strong global model to excel in the generic FL setup.

Some other approaches are based on mixture models. \citep{peterson2019private,deng2020adaptive,mansour2020three,agarwal2020federated,zec2020federated} (separately) learned global and personalized models and performed a mixture of them for prediction. \citep{reisser2021federated} learned a sets of expert models and used them to construct personalized models.
Meta-learning is also applied to learn a good initialized model that can be adapted to each client with a few steps of local training~\citep{Khodak2019AdaptiveGM,Chen2018FederatedMW,fallah2020personalized,Jiang2019ImprovingFL}.

Instead of designing specific algorithms for personalized FL,  \citep{yu2020salvaging,Wang2019FederatedEO} showed that performing post-processing (\eg, fine-tuning) to a generic FL model (\eg, $\bar{\vw}$ learned by \FedAvg) already leads to promising personalized accuracy. In this work, we further showed that, the \emph{local models} $\vw_m$ learned in \FedAvg and other generic FL algorithms are indeed strong personalized models.

\emph{We note that, while many personalized FL algorithms also produce a global model, it is mainly used to regularize or construct personalized models but not for evaluation in the generic setup. In contrast, we learn models to excel in both the setups via a single framework without sacrificing
either of them.} 

\pFedHN~\citep{shamsian2021personalized} also applies hypernetworks~\citep{ha2016hypernetworks} but for a very different purpose from \FedRoD. Specifically, \pFedHN learns a hypernetwork at the server to aggregate clients' model updates and produce their entire models for the next round. In contrast, we learn the hypernetwork locally to construct the personalized predictors, not the entire models, for fast adaptation to clients.

\subsection{Averaging model weights as a regularizer}
In~\autoref{ss_strong_reg}, we demonstrate that taking the average over model weights indeed acts as a regularizer for \emph{local models} to improve their individual \emph{personalized} performance.

In more traditional machine learning, the regularization effects of averaging multiple independently-trained models have been observed in some techniques like bagging~\citep{poggio2002bagging,skurichina1998bagging}. Indeed, in several recent works of FL~\citep{lin2020ensemble,he2020group,zhou2020distilled,chen2021fedbe}, the authors replaced weight average by bagging/model ensemble to improve the \emph{generic} performance on the global test set. That is, they found that performing the model ensemble over clients’ models can yield more robust predictions on the global test set than the global model, which is generated by averaging the client models’ weights.

Here, we however study a different regularization effect, in personalized FL on local test sets. As reviewed in~\autoref{ss_back_G}, personalized FL algorithms often impose a regularizer on the local/personalized models to overcome the fact that clients usually have limited data (please see~\autoref{eq:P-obj} and~\autoref{eq_pfl_local} and the surrounding text). What we claim is that even without such an explicit regularizer, the model weight average before local training (\autoref{eq_gfl_local}) already serves as an implicit regularizer to the local models for their individual personalized performance, as we discussed in~\autoref{ss_strong_reg} (\autoref{fig:check}) and empirically verified in~\autoref{sup_subs_local_meta} and~\autoref{fig:Ditto_grad}.

\subsection{Systematic overhead}
\FedRoD has similar computation cost, communication size, and number of parameters as \FedAvg. We discuss the difference between \FedRoD and existing generic FL methods from a system view. \FedProx~\citep{li2020federated} proposes a proximal term to prevent client from diverging from the server model, which is more robust to the heterogeneous system. \SCAFFOLD~\citep{karimireddy2020scaffold} imposes a gradient correction during client training. Maintaining such a correction term, however, doubles the size of communication. \FedDyn~\citep{feddyn} resolves the communication cost issue by introducing a novel dynamic regularization. However, it requires all users to maintain their previous models locally throughout the FL process, which is not desired when users have memory and synchronization constraints. 

\subsection{Class-imbalanced learning} 
Class-imbalanced learning attracts increasing attention for two reasons. First, models trained under this scenario using empirical risk minimization perform poorly on minor classes of scarce training data. Second, many real-world data sets are class-imbalanced by nature \citep{van2018inaturalist,gupta2019lvis,van2017devil}. In this paper, we employ a mainstream approach, cost-sensitive learning \citep{ye2020identifying,cao2019learning,ren2020balanced,li2020overcoming,tan2020equalization}, which adjusts the training objective to reflect class imbalance so as to train a model that is less biased toward major classes.

\subsection{Zero-shot learning} Our design choice of parameterizing the personalized prediction head with clients' class distributions is reminiscent of zero-shot learning \citep{changpinyo2016synthesized,changpinyo2020classifier,xian2018zero,changpinyo2017predicting,lampert2013attribute}, whose goal is to build an object classifier based on its semantic representation. The key difference is that we build an entire fully-connected layer for FL, not just a single class vector. We employ hypernetworks \citep{ha2016hypernetworks} for efficient parameterization.

\section{Additional Details of \FedRoD}
\label{suppl-sec:full}

\subsection{Additional background (cf. \autoref{ss_back_G} of the main paper)}
In the \emph{generic} federated learning (FL) setting, the goal is to construct a single ``global'' model that can perform well for test data from all the clients. Let $\vw$ denote the parameters of the model, for a classification problem whose label space is $\C$, a commonly used loss is the cross entropy,
\begin{align}
\ell(\vx, y; \vw) =  -\log\frac{\exp(g_y(\vx; \vw))}{\sum_{c\in\C} \exp(g_c(\vx; \vw))}, 
\label{sup_eq:CE}
\end{align}
where $g_c(\vx; \vw)$ is the model's output logit for class $c$.

We note that, the concepts of global vs. local models and generic vs. personalized models should not be confused. No matter which task (generic or personalized) an FL algorithm focuses on, as long as it has the local training step, it generates local models; as long as it has the global aggregation step (of the entire model), it generates a global model. For instance, \FedAvg~\citep{mcmahan2017communication} aims for generic FL but it creates both the global and local models.

\subsection{Overview of \FedRoD}

For generic predictions, \FedRoD performs feature extraction $\vz = f(\vx;\vtheta)$, followed by $h^G(\vz; \vpsi)$. For personalized predictions, \FedRoD performs $\vz=f(\vx;\vtheta)$, followed by $h^G(\vz; \vpsi)+h^P(\vz; \vphi_m)$. The element-wise addition is performed at the \emph{logit} level. That is, $g_c(\vx; \vw)$ in~\autoref{sup_eq:CE} can be re-written as 
\begin{align}
g_c(\vx; \{\vtheta, \vpsi, \vphi_m\})=
    \begin{cases}
        h_c^G(\vz; \vpsi)  & \textbf{Generic model}, \\
        h_c^G(\vz; \vpsi) + h_c^P(\vz; \vphi_m) & \textbf{Personalized model},\\
    \end{cases} \label{sup_eq_gm}
\end{align}
where $\vz = f(\vx;\vtheta)$ is the extracted feature.

The overall training process of \FedRoD iterates between the local training and global aggregation steps. In local training, \FedRoD aims to minimize the following objective 
\begin{align}
\sL_m^{\textcolor{red}{BR}}(\{\vtheta, \vpsi\}) + \sL_m(\{\vtheta, \vpsi, \vphi_m\}).
\label{sup_eq_fedrod_learn}
\end{align}
The empirical risk $\sL_m(\vw_m = \{\vtheta, \vpsi, \vphi_m\})$ is defined as $\frac{1}{|\sD_m|} \sum_{i} \ell(\vx_i, y_i; \vw_m)$, where $\sD_m = \{(\vx_i, y_i)\}_{i=1}^{|\sD_m|}$ is the training data of client $m$. We will introduce more options of the balanced risk $\sL_m^{\textcolor{red}{BR}}$ in \autoref{suppl_s_BSM}. 
We optimize \autoref{sup_eq_fedrod_learn} via stochastic gradient descent (SGD). We updates $\vtheta$, $\vpsi$, and $\vphi_m$ in a single forward-backward pass, which consumes almost the same computation cost as \FedAvg. For $\sL_m(\{\vtheta, \vpsi, \vphi_m\})$, we do not derive gradients w.r.t. $\vtheta$ and  $\vpsi$.

We emphasize that, according to \autoref{ss_decouple} of the main paper, the finally learned parameters of \FedRoD (linear) are $\bar{\vtheta}$, $\bar{\vpsi}$, and $\{\vphi^\star_m\}_{m=1}^M$. We then plug them into \autoref{sup_eq_gm} for predictions.

In \autoref{alg:linear} and \autoref{alg:hyper}, we provide pseudocode of our \FedRoD algorithm.

\begin{algorithm}[H]
\footnotesize
\SetAlgoLined
\caption{\FedRoD (linear) \hspace{2pt} {\small(\textbf{Fed}erated \textbf{Ro}bust \textbf{D}ecoupling)}}
\label{alg:linear}
\SetKwInOut{Input}{Input}
\SetKwInOut{SInput}{Server input}
\SetKwInOut{CInput}{Client $m$'s input}
\SetKwInOut{SOutput}{Server output}
\SetKwInOut{COutput}{Client $m$'s output}
\SInput{initial global model parameter $\bar{\vtheta}$ and $\bar{\vpsi}$;}
\CInput{initial local model parameter ${\vphi^\star_m}$, local step size $\eta$, local labeled data $\mathcal{D}_m$;
}
\For{$r\leftarrow 1$ \KwTo $R$}{
{\textbf{Sample} clients $\mathcal{S}\subseteq\{1,\cdots,N\}$};\\
{\textbf{Communicate} $\bar{\vtheta}$ and $\bar{\vpsi}$ to all clients $m\in \mathcal{S}$;}\\
\For{each client $m\in \mathcal{S}$ in parallel}{
{\textbf{Initialize} $\vtheta\leftarrow\bar{\vtheta}$, $\vpsi\leftarrow\bar{\vpsi}$, and $\vphi_m\leftarrow{\vphi^\star_m}$;}\\
{$\{\vtheta_m^\star, \vpsi_m^\star, \vphi_m^\star\} \leftarrow \textbf{Client local training}(\{\vtheta, \vpsi, \vphi_m\}, \mathcal{D}_m, \eta)$}; \hfill {[\autoref{eq_ours_local_g} and \autoref{eq_ours_local_p}]} \\
{\textbf{Communicate} $\vtheta_m^\star$ and $\vpsi_m^\star$ to the server;}\\
}
\textbf{Construct} $\bar \vtheta =\sum_{m\in \mathcal{S}} \frac{|\mathcal{D}_m|}{\sum_{m'\in \mathcal{S}}|\mathcal{D}_{m'}|} \vtheta^\star_m$;\\
\textbf{Construct} $\bar \vpsi =\sum_{m\in \mathcal{S}} \frac{|\mathcal{D}_m|}{\sum_{m'\in \mathcal{S}}|\mathcal{D}_{m'}|} \vpsi^\star_m$;\\
}
\SOutput{$\bar{\vtheta}$ and $\bar{\vpsi}$;}
\COutput{$\{\vtheta_m^\star, \vpsi_m^\star, \vphi_m^\star\}$.}
\end{algorithm}

\begin{algorithm}[H]
\footnotesize
\SetAlgoLined
\caption{\FedRoD (hyper) \hspace{2pt} {\small(\textbf{Fed}erated \textbf{Ro}bust \textbf{D}ecoupling)}}
\label{alg:hyper}
\SetKwInOut{Input}{Input}
\SetKwInOut{SInput}{Server input}
\SetKwInOut{CInput}{Client $m$'s input}
\SetKwInOut{SOutput}{Server output}
\SetKwInOut{COutput}{Client $m$'s output}
\SInput{initial global model parameter $\bar{\vtheta}$, $\bar{\vpsi}$, and $\bar{\vnu}$;}
\CInput{local step size $\eta$, local labeled data $\mathcal{D}_m$;
}
\For{$r\leftarrow 1$ \KwTo $R$}{
{\textbf{Sample} clients $\mathcal{S}\subseteq\{1,\cdots,N\}$};\\
{\textbf{Communicate} $\bar{\vtheta}$, $\bar{\vpsi}$, and $\bar{\vnu}$ to all clients $m\in \mathcal{S}$;}\\
\For{each client $m\in \mathcal{S}$ in parallel}{
{\textbf{Initialize} $\vtheta\leftarrow\bar{\vtheta}$, $\vpsi\leftarrow\bar{\vpsi}$, and $\vnu\leftarrow{\bar{\vnu}}$;}\\
{$\{\vtheta_m^\star, \vpsi_m^\star, \vnu_m^\star\} \leftarrow \textbf{Client local training}(\{\vtheta, \vpsi, \vnu\}, \mathcal{D}_m, \eta)$}; \hfill {[\autoref{eq_meta_hyper}]} \\
{\textbf{Communicate} $\vtheta_m^\star$, $\vpsi_m^\star$, and $\vnu_m^\star$ to the server;}\\
}
\textbf{Construct} $\bar \vtheta =\sum_{m\in \mathcal{S}} \frac{|\mathcal{D}_m|}{\sum_{m'\in \mathcal{S}}|\mathcal{D}_{m'}|} \vtheta^\star_m$;\\
\textbf{Construct} $\bar \vpsi =\sum_{m\in \mathcal{S}} \frac{|\mathcal{D}_m|}{\sum_{m'\in \mathcal{S}}|\mathcal{D}_{m'}|} \vpsi^\star_m$;\\
\textbf{Construct} $\bar \vnu =\sum_{m\in \mathcal{S}} \frac{|\mathcal{D}_m|}{\sum_{m'\in \mathcal{S}}|\mathcal{D}_{m'}|} \vnu^\star_m$;\\
}
\SOutput{$\bar{\vtheta}$, $\bar{\vpsi}$, and $\bar{\vnu}$ (for personalized model generation).}
\COutput{$\{\vtheta_m^\star, \vpsi_m^\star, \vnu_m^\star\}$.}
\end{algorithm}

\subsection{Balanced risk minimization (BRM)}
\label{suppl_s_BSM}

To learn a generic model, standard federated learning (\eg, \FedAvg ~\citep{mcmahan2017communication}) aims to optimize $\sL(\vw)$ in \autoref{eq:obj} of the main paper. In theory, the overall objective $\sL(\vw)$ is equal to the expected objective $\mathbb{E}[\sL_m(\vw)]$ for client $m$, if client $m$'s data $\mathcal{D}_m$ are IID partitioned from $\mathcal{D}$. Here, the expectation is 
over different $\mathcal{D}_m$ partitioned from $\mathcal{D}$.
In reality, $\sL_m$ could diverge from $\sL$ due to non-IID partitions of the aggregated data $\mathcal{D}$ into clients' data. That is, $\mathbb{E}[\sL_m(\vw)] \neq \mathbb{E}[\sL_{m'}(\vw)] \neq \sL(\vw)$. We mitigate the non-IID situation by directly adjusting $\sL_m$ such that $\mathbb{E}[\sL_m(\vw)] \approx \mathbb{E}[\sL_{m'}(\vw)] \approx \sL(\vw)$. 

Essentially, $\sL_m(\vw)$ is the client's empirical risk, which could be different among clients if their class distribution $\sP_m(y)$ are different. We, therefore, propose to turn the empirical risk $\sL_m$ into a class-balanced risk $\sL_m^{\textcolor{red}{BR}}$ by replacing $\ell$ in \autoref{sup_eq:CE} with a class-balanced loss \citep{ren2020balanced,cao2019learning,cui2019class,ye2020identifying,khan2017cost,kang2019decoupling}. 
The class-balanced loss attempts to make the learned model robust to different training class distributions, such that the learned model can perform well for all the test classes. In other words, the class-balanced loss is designed with an implicit assumption that the test data will be class-balanced, even though the training data may not be.
\autoref{sup_tbl:summary_loss} summarizes some popular class-balanced losses. We also include some extensions with meta-learning. See~\autoref{sup_subsec_meta}.

One may wonder what if the global distribution is class-imbalanced? Will BRM still be beneficial to FL? In~\autoref{sup_subs_imb}, we perform experiments to show that \FedRoD with BRM can still improve on \FedAvg since BRM seeks to learn every class well. Even though the global distribution might be skewed, BRM provides a novel alternative to mitigate the non-IID problem by making every client optimize a more consistent objective as we discussed above. Designing better losses for BRM in FL will be interesting future work. 

\begin{table*}[t] 
    \small
	\centering
	\caption{\small Balanced risk and loss functions. We ignore the normalization in $\sL_m(\vw)$. \textcolor{red}{Red} highlights the modifications by the balanced losses. \textcolor{blue}{Blue} highlights the terms learned with meta-learning (see~\autoref{sup_subsec_meta}).} 
	\setlength{\tabcolsep}{2pt}
	\renewcommand{\arraystretch}{1}
	\begin{tabular}{l|cc}
    \midrule
    {Method} & $\ell(\vx, y; \vw)$ & $\sL_m$ or $\sL_m^{\textcolor{red}{BR}}$\\
    \midrule
    Cross entropy & $-\log\frac{\exp(g_y(\vx; \vw))}{\sum_{c\in\C} \exp(g_c(\vx; \vw))}$ & $\sum_i \ell(\vx_i, y_i; \vw)$\\
    \midrule
    \midrule
    IR~\citep{hsu2020federated} &   $-\log\frac{\exp(g_y(\vx; \vw))}{\sum_{c\in\C} \exp(g_c(\vx; \vw))}$ & $\sum_{i} {\color{red}\frac{\sum_{c\in\C}N_{m,c}}{N_{m,y_i}}}\ell(\vx_i, y_i; \vw)$\\\\
    \midrule
    LDAM~\citep{cao2019learning}\\ ($\textcolor{red}{\gamma}$ tuned with validation) & $-\log\frac{\exp( g_y(\vx; \vw) - \textcolor{red}{\gamma N^{-\frac{1}{4}}_{m,y}})}{\sum_{c\in\C, c\neq y} \exp(g_c(\vx; \vw)) + \exp(g_y(\vx; \vw)- \textcolor{red}{\gamma N^{-\frac{1}{4}}_{m,y}})}$ & $\sum_i \ell(\vx_i, y_i; \vw)$\\\\
    \midrule
    CDT~\citep{ye2020identifying}\\ ($\textcolor{red}{\gamma}$ tuned with validation) &  $-\log\frac{\exp(\textcolor{red}{(\frac{N_{m,y}}{N_{m,\text{max}}})^\gamma} g_y(\vx; \vw))}{\sum_{c\in\C} \exp(\textcolor{red}{(\frac{N_{m,c}}{N_{m,\text{max}}})^\gamma} g_c(\vx; \vw))}$ & $\sum_i \ell(\vx_i, y_i; \vw)$\\\\
    \midrule
    BSM~\citep{ren2020balanced}\\ ($\textcolor{red}{\gamma}=1$ fixed) & $-\log\frac{{\color{red}N^{\gamma}_{m,y}}\exp(g_y(\vx; \vw))}{\sum_{c\in\C} {\color{red}N^{\gamma}_{m,c}} \exp(g_c(\vx; \vw))}$ & $\sum_i \ell(\vx_i, y_i; \vw)$\\\\
    \midrule
    \midrule
    Meta-BSM \\($\textcolor{red}{\gamma}=1$ fixed, \\$\textcolor{blue}{q_{m, y_i}}$ meta-learned) &  $-\log\frac{{\color{red}N^{\gamma}_{m,y}}\exp(g_y(\vx; \vw))}{\sum_{c\in\C} {\color{red}N^{\gamma}_{m,c}} \exp(g_c(\vx; \vw))}$ & $\sum_i {\color{blue}q_{m, y_i}}\ell(\vx_i, y_i; \vw)$\\\\
    \midrule
    Meta-BSM \\($\textcolor{blue}{\gamma_m, q_{m, y_i}}$ meta-learned) & $-\log\frac{{\color{red}N_{m,y}}^{\color{blue}\gamma_m}\exp(g_y(\vx; \vw))}{\sum_{c\in\C} {\color{red}N}_{\color{red}{m,c}}^{\color{blue}\gamma_m} \exp(g_c(\vx; \vw))}$ & $\sum_i {\color{blue}q_{m, y_i}}\ell(\vx_i, y_i; \vw)$\\\\
    \bottomrule
	\end{tabular}
	\label{sup_tbl:summary_loss}
\end{table*}

\subsection{On federated learning for the personalized head with hypernetworks} \label{sup_subsec_hype}

\begin{table}[t] 
    \footnotesize
	\centering
	\caption{\small $\#$ of parameters in ConvNets for EMNIST/FMNIST and CIFAR-10/100} 
	\setlength{\tabcolsep}{4pt}
	\renewcommand{\arraystretch}{0.5}
	\begin{tabular}{l|ccc}
	\toprule
	Module & EMNIST/FMNIST & CIFAR-10 & CIFAR-100\\
	\midrule
	Feature extractor & 92,646 & 1,025,610 & 1,025,610\\
	Generic head & 500 & 640 & 6400\\
    Total & 93,146 & 1,026,250 & 1,032,010\\
    \midrule
    Hypernetworks & 8,160 ($+8.8\%$) & 20,800 ($+2.0\%$) & 104,000 ($+10.0\%$)\\
    \bottomrule
	\end{tabular}
	\label{tbl:para}
\end{table}

One drawback of existing personalized methods is that the personalized models are only available for clients involved in training or with sufficient training data. When new clients arrive in testing, is it possible for the federated system to provide corresponding personalized models? 

To this end, instead of learning a specific prediction head $\vphi_m$ for each client $m$, we propose to learn a meta-model $H^P(\vct{a}_m; \vnu)$ with a shared meta-parameter $\vnu$. The input to $H^P$ is a vector $\vct{a}_m \in \R^{|\C|}$, which records the proportion of class $c\in \C$ in client $m$'s data. The output of $H^P$ is $\vphi_m$ for $h^P$. In other words, $H^P$ can adaptively output personalized prediction heads for clients given their local class distributions $\vct{a}_m$.

We implement the meta-model $H^P$ by a hypernetwork \citep{ha2016hypernetworks}, which can be seen as a lightweight classifier generator given $\vct{a}_m$. This lightweight hypernetwork not only enables clients to collaboratively learn a module that can generate customized models, but also allows any (future) clients to immediately generate their own personalized predictors given their local class distribution $\vct{a}_m$ as input, even without training. We construct the hypernetwork by two fully-connected (FC) layers (with a ReLU nonlinear layer in between). \autoref{tbl:para} summarizes the number of parameters of each part in \FedRoD. Hypernetworks add only a small overhead to the original model.

\subsection{Extension with meta-learning for the improved BSM loss}
\label{sup_subsec_meta}
\FedRoD incorporates a balanced loss to learn the generic model. Here we study a more advanced way to derive such balanced loss with meta-learning. Inspired by~\citep{ren2018learning,Shu2019Meta} and the FL scenario proposed by~\citep{zhao2018federated}, we seek to combine the BSM loss and  re-weighting as $\sum_i {\color{blue}q_{m, y_i}}\ell^\text{BSM}(\vx_i, y_i; \vw)$, where ${\color{blue}q_{m, y_i}}$ is meta-learned with a small balanced meta dataset $\sD_{\text{meta}}$ provided by the server. (See \autoref{sup_tbl:summary_loss} for a comparison.) The $\sD_{\text{meta}}$ should have a similar distribution to the future test data. We implement this idea with the Meta-Weight Net (MWNet)~\citep{Shu2019Meta} with learnable parameter $\vzeta$. 

In addition, we notice that the original BSM loss $\ell^{\text{BSM}}_{\gamma} = -\log\frac{{\color{red}N^{\gamma}_{m,y}}\exp(g_y(\vx; \vw))}{\sum_{c\in\C} {\color{red}N^{\gamma}_{m,c}} \exp(g_c(\vx; \vw))}$ has a hyperparameter $\gamma$ which is set to be $1$ via validation~\citep{ren2020balanced}. However, in federated learning it can be hard to tune such a hyperparameter due to the large number of non-IID clients. Therefore, we propose to learn a client-specific $\color{blue}\gamma_m$ with meta-learning for $\ell^{\text{BSM}}_{\color{blue}\gamma_m}$. More specifically, given a meta-learning rate $\eta$, the meta-learning process involves the following iterative steps:
\begin{enumerate}
\item Compute the Meta-BSM loss with a mini-batch $B\sim \sD_m$; \ie, $\forall (\vx,y) \in B$, compute $\ell^{\text{BSM}}_{\color{blue}\gamma_m}(\vx, y; \vw_m)$.
\item Predict the example weights with $\textcolor{blue}{q_{m, y}} = \text{MWNet}(\ell^{\text{BSM}}_{\color{blue}\gamma_m}(\vx, y; \vw_m); \vzeta_m)$, $\forall (\vx,y) \in B$.
\item Re-weight the Meta-BSM loss: $\sL_{m,B}^{\text{BR}}(\vw_m) = \sum_{(\vx, y)\in B} \textcolor{blue}{q_{m, y_i}}\ell^{\text{BSM}}_{\color{blue}\gamma_m}(\vx, y; \vw_m)$, and perform one step of gradient descent to create a duplicated model $\tilde \vw_m =  \vw_m - \eta\nabla_{\vw_m}\sL_{m,B}^{\text{BR}}$.
\item Computes the loss on the meta dataset $\sD_{\text{meta}}$ using the duplicated model: $\sL_{m,\sD_{\text{meta}}}^{\text{BR}}(\tilde \vw_m) = \sum_{(\vx, y)\in \sD_{\text{meta}}} \textcolor{blue}{q_{m,y}}\ell^{\text{BSM}}_{\color{blue}\gamma_m}(\vx, y, \tilde \vw_m)$, followed by updating $\textcolor{blue}{\gamma_m} \leftarrow \textcolor{blue}{\gamma_m} - \eta\nabla_{\color{blue}\gamma_m}\sL_{m,\sD_{\text{meta}}}^{\text{BR}}$ and $\vzeta_m \leftarrow \vzeta_m - \eta\nabla_{\vzeta_m}\sL_{m,\sD_{\text{meta}}}^{\text{BR}}$.
\item Update the model: $\vw_m \leftarrow \vw_m - \eta\nabla_{\vw_m}\sL_{m,B}^{\text{BR}}(\vw_m)$.
\end{enumerate}

Throughout the federated learning process, $\color{blue}\gamma_m$ and $\color{blue}q_{m,y}$ are dynamically learned with meta-learning for different clients and rounds.

\paragraph{Results of \FedRoD with Meta-BSM} \label{ss_meta_results}
We sample \textbf{10} images for each class (only $0.2\%$ of the overall training set) from the training set as the meta set. We compare to~\citep{zhao2018federated} that concatenates the meta set to clients' local data. The results in \autoref{tbl:main_imb} and \autoref{sup_tbl:main_bal} are encouraging. With a very small meta set, \FedRoD outperforms \citep{zhao2018federated} by $1\%$ to $14\%$ on accuracy across different settings, validating the importance of balanced losses and how to set them up dynamically via meta-learning.  

\subsection{Connection to Invariant Risk Minimization Games (IRMG)} \label{suppl-subsec:irmg}
\FedRoD is inspired by a recently proposed machine learning framework Invariant Risk Minimization (IRM)~\citep{arjovsky2019invariant} and its extension Invariant Risk Minimization Games (IRMG)~\citep{ahuja2020invariant}.

Suppose that the whole dataset is collected from many environments, where data from each environment is associated with its characteristic, IRM introduces the concept of learning an invariant predictor.
(Note that, in IRM the learner can access data from all the environments; thus, it is not for an FL setting.) Given the training data partition, IRM aims to learn an invariant feature extractor $\vz = f(\vx;\vtheta)$ and a classifier $h(\vz; \vpsi)$ that achieves the minimum risk for all the environments. 

The concept of \emph{environments} can be connected to clients' private local data in FL which are often non-IID. That is, given $M$ environments, we can re-write IRM in a similar expression to \autoref{eq_fedrod} in the main paper
\begin{align}
&\min_{\vtheta, \vpsi}~\sL^{\text{IRM}}(\vtheta, \vpsi) = \sum_{m=1}^M \sL_m(\vtheta, \vpsi),\\
&\text{s.t}\quad \vpsi \in  \argmin_{\vpsi'} \sL_m(\vtheta, \vpsi'), \forall m \in [M].
\label{sup_eq:irm}
\end{align}

Unfortunately, IRM is intractable to solve in practice given the constraint that every environment relies on the same parameters~\citep{ahuja2020invariant}. IRMG relaxes it by reformulating the classifier $\vpsi$ as an ensemble of environment-specific classifiers (by averaging over model weights) $\bar \vphi = \frac{1}{M}\sum_m \vphi_m$:
\begin{align}
&\min_{\vtheta, \bar \vphi}~\sL^{\text{IRMG}}(\vtheta, \bar \vphi) = \sum_{m=1}^M \sL_m(\vtheta, \bar \vphi), \label{sup_eq:irmg-server}\\
&\text{s.t}\quad \vphi_m \in  \argmin_{\vphi_{m'=m}} \sL_m(\vtheta, \{\vphi_{m'}\}_{m'=1}^{M}), \forall m \in [M].
\label{sup_eq:irmg-client}
\end{align}
IRMG is proved to optimize the same invariant predictor of IRM when it converges to the equilibrium in game theory, and it holds for a large class of non-linear classifiers. IRMG is solved through iterative optimization: (1) training the feature extractor $\vtheta$ with centralized data (\ie, aggregated data from all environments), (2) training the environment-specific classifiers $\vphi_m$ on the data of each environment $\sD_m$, and (3) updating the main classifier through weight averaging $\bar \vphi = \frac{1}{M}\sum_m \vphi_m$.   

We highlight the similarity between IRMG and \FedRoD: both are training a strong generic feature extractor and a set of personalized classifiers. For predictions on data of client (environment) $m$ in~\autoref{sup_eq:irmg-client}, IRMG uses $\hat y = \frac{1}{M}(\phi_m^\top \vz + \sum_{m' \neq m}\vphi_{m'}^\top \vz)$; \FedRoD's personalized model is $\hat y = h^G(\vz; \vpsi)+h^P(\vz; \vphi_m)$. 
We can connect IRMG to \FedRoD by re-writing its prediction as $h^G(\vz; \bar{\vphi}) := \bar{\vphi}^\top \vz=\frac{1}{M}\sum_{m} {\vphi'}_{m}^\top \vz$ and $h^P(\vz; \vphi_m) := \frac{1}{M} (\vphi_m^\top \vz-{\vphi'_m}^\top \vz)$, where ${\vphi'_m}$ is the client $m$'s model in the previous round/iteration of learning.  

IRMG can not be applied directly to federated learning for the following reasons.
First, centralized training of the feature extractor is intractable since clients' data are not allowed to be aggregated to the server. Second, to perform the iterative optimization of IRMG, the clients are required to communicate every step, which is not feasible in FL due to communication constraints.

\section{Implementation Details}
\label{suppl-sec:exp_s}

\paragraph{Implementation.}
We adopt ConvNet~\citep{lecun1998gradient} following the existing works \citep{mcmahan2017communication,feddyn,tfconvnet16}. 
For EMNIST/FMNIST, it contains $2$ Conv layers and $2$ FC layers. The Conv layers have $32$ and $64$ channels, respectively. The FC layers are with $50$ neurons as the hidden size and $10$ neurons for $10$ classes as outputs, respectively. For CIFAR-10/100, it contains $3$ Conv layers and $2$ FC layers. The Conv layers have $32$, $64$, and $64$ channels, respectively. The FC layers are with $64$ neurons as the hidden size and $10/100$ neurons for $10/100$ classes as outputs, respectively. To implement hypernetworks in \FedRoD, we use a simple 2-FC ReLU network with hidden size $16$ for EMNIST/FMNIST/CIFAR-100 and $32$ for CIFAR-10.  

We use standard pre-processing, where EMNIST/FMNIST and CIFAR-10/100 images are normalized. EMNIST/FMNIST is trained without augmentation. The $32 \times 32$ CIFAR-10/100 images are padded $2$ pixels each side, randomly flipped horizontally, and then randomly cropped back to $32 \times 32$. 

We train every method for $100$ rounds. We initialize the model weights from normal distributions.
As mentioned in \citep{li2020convergence}, the local learning rate must decay along the communication rounds. We initialize it with $0.01$ and decay it by $0.99$ every round, similar to~\citep{feddyn}. Throughout the experiments, we use the \SGD optimizer with weight decay $1\mathrm{e}{-5}$ and a $0.9$ momentum. The mini-batch size is $40$ ($16$ for EMNIST). In each round, clients perform local training for $5$ epochs. We report the mean over five times of experiments with different random seeds.

For \FedProx~\citep{li2020federated}, the strength of regularization $\lambda$ is selected from $[\mathrm{1e}{-2}, \mathrm{1e}{-3}, \mathrm{1e}{-4}]$. For \FedDyn \citep{feddyn}, the strength of regularization $\lambda$ is selected from $[\mathrm{1e}{-1}, \mathrm{1e}{-2}, \mathrm{1e}{-3}]$ as suggested in \citep{feddyn}. For \PerFedAvg~\citep{fallah2020personalized}, the meta-learning rate $\hat \beta$ is selected from $[\mathrm{1e}{-2}, \mathrm{1e}{-3}, \mathrm{1e}{-4}]$. For \pFedMe~\citep{dinh2020personalized}, the strength of regularization $\lambda$ is selected from $[15, 20, 30]$. \FedRoD introduces no extra hyperparameters on top of \FedAvg. 

For the generic and personalized heads of \FedRoD, we study using $1\sim4$ FC layers but do not see a notable gain by using more layers. We attribute
this to the well-learned generic features. Thus, for all our experiments on \FedRoD, we use a single FC layer for each head.

We run our experiments on four GeForce RTX 2080 Ti GPUs with Intel i9-9960X CPUs.

\paragraph{Evaluation.}
Both datasets and the non-IID Dirichlet simulation are widely studied and used in literature~\citep{hsu2019measuring,lin2020ensemble,feddyn}.
We use the standard balanced test set $\sD_{\text{test}}$ for evaluation on generic FL (G-FL):
\begin{align}
    \textbf{G-FL}\hspace{4pt}\text{accuracy}:  \hspace{4pt}\frac{1}{|\sD_{\text{test}}|}\sum_i \textbf{1}(y_i = \hat y_{i, G}),
\end{align}
where $\hat y_{i, G}$ here is the predicted label (\ie, $\argmax$ over the logits). For evaluation on personalized FL (P-FL), we still apply $\sD_{\text{test}}$ but weight instances w.r.t. each client's class distribution:
  \begin{align}
    \textbf{P-FL}\hspace{4pt}\text{accuracy}:  \hspace{4pt}\frac{1}{M}\sum_m\frac{\sum_i\sP_m(y_i)\textbf{1}(y_i = \hat y_{i, P})}{\sum_i\sP_m(y_i)}.
\end{align}
We do so instead of separating $\sD_{\text{test}}$ into separate clients' test sets in order to avoid the variance caused by how we split test data (except the EMNIST dataset that each client has its own test set with the writer's styles). What we compute is essentially the expectation over the splits. We have verified that the difference of the two evaluation methods is negligible. 
  
In \autoref{tbl:main_bal} of the main paper and some other tables in the appendix, we evaluate G-FL by an FL algorithm's generic (usually the global) model, denoted as \textbf{GM}. We evaluate P-FL by an FL algorithm's personalized models (or local models of a G-FL algorithm), denoted as \textbf{PM}. For P-FL, we also report the generic model's accuracy following the literature to demonstrate the difference.

Due to the space limit of the main paper, we provide the standard deviations of the results of \autoref{tbl:main_bal} in \autoref{sup-tbl:main_var1}, \autoref{sup-tbl:main_var2}, and \autoref{sup-tbl:main_var3} here. 

\begin{table*}[t!] 
    \scriptsize
	\centering
	\caption{\small EMNIST and FMNIST results in G-FL accuracy and P-FL accuracy ($\%$). $\star$: methods with no G-FL models and we combine their P-FL models. $\mathsection$: official implementation.} 
	\vspace{-5pt}
	\setlength{\tabcolsep}{3pt}
	\renewcommand{\arraystretch}{0.5}
	\begin{tabular}{l|ccc|ccc|ccc|ccc|ccc}
	\toprule {Dataset} &
	\multicolumn{3}{c|}{EMNIST}&
	\multicolumn{6}{c|}{FMNIST}\\
	\midrule {Non-IID} &
	\multicolumn{3}{c|}{Writers
} & \multicolumn{3}{c|}{Dir(0.1)
} & \multicolumn{3}{c}{Dir(0.3)} \\
    \midrule
    {Test Set} & G-FL & \multicolumn{2}{|c|}{P-FL} & G-FL & \multicolumn{2}{|c|}{P-FL} & G-FL & \multicolumn{2}{|c|}{P-FL} \\
    \midrule
    
    {Method / Model} &
    \multicolumn{1}{c|}{GM} & GM & PM & \multicolumn{1}{c|}{GM} & GM & PM & \multicolumn{1}{c|}{GM} & GM & PM  \\
    \midrule
    \FedAvg& 97.0$\pm$0.05 & 96.9$\pm$0.05 & 97.2$\pm$0.06 & 81.1$\pm$0.12 & 81.0$\pm$0.14 & 91.5$\pm$0.14 & 83.4$\pm$0.15 &83.2$\pm$0.15 & 90.5$\pm$0.21 \\ 
    \FedProx& 97.0$\pm$0.05 & 97.0$\pm$0.05 & 97.0$\pm$0.05 & 82.2$\pm$0.15 & 82.3$\pm$0.13& 91.4$\pm$0.10 & 84.5$\pm$0.14 & 84.5$\pm$0.17 & 89.7$\pm$0.19 \\
    \SCAFFOLD& 97.1$\pm$0.11 & 97.0$\pm$0.12 & 97.1$\pm$0.09 & 83.1$\pm$0.25 & 83.0$\pm$0.30 & 89.0$\pm$0.32 & 85.1$\pm$0.27 & 85.0$\pm$0.29 & 90.4$\pm$0.34\\
    \FedDyn$\mathsection$& 97.3$\pm$0.12 & 97.3$\pm$0.10 &97.3$\pm$0.10 & 83.2$\pm$0.15 & 83.2$\pm$0.16 &90.7$\pm$0.20 & 86.1$\pm$0.18 & 86.1$\pm$0.17 & 91.5$\pm$0.19\\
    \midrule
    \MOCHA$^\star$ & 75.4$\pm$0.85 & 75.0$\pm$0.78 & 85.6$\pm$0.77& 36.1$\pm$0.65 & 36.0$\pm$0.66 & 87.3$\pm$0.75 & 53.1$\pm$0.70 & 53.4$\pm$0.69 & 78.3$\pm$0.80 \\
    \LGFedAvg$^\star\mathsection$ & 80.1$\pm$0.34 & 80.0$\pm$0.24 & 95.6$\pm$0.15 &54.8$\pm$0.41 & 54.5$\pm$0.44 & 89.5$\pm$0.64 & 66.8$\pm$0.40 & 66.8$\pm$0.42 & 84.4$\pm$0.55 \\
    \FedPer$^\star$& 93.3$\pm$0.14 & 93.1$\pm$0.20 & 97.2$\pm$0.11 & 74.5$\pm$0.24&74.4$\pm$0.25&91.3$\pm$0.48 & 79.9$\pm$0.20&79.9$\pm$0.22&90.4$\pm$0.41\\
    \PerFedAvg & 95.1$\pm$0.24 & - & 97.0$\pm$0.14 & 80.5$\pm$0.60 & - & 82.8$\pm$1.20 & 84.1$\pm$0.75 & - & 86.7$\pm$0.99 \\
    \pFedMe$\mathsection$ & 96.3$\pm$0.11 & 96.0$\pm$0.10 & 97.1$\pm$0.11 & 76.7$\pm$0.33 & 76.7$\pm$0.35 & 83.4$\pm$0.41 & 79.0$\pm$0.35 & 79.0$\pm$0.35 & 83.4$\pm$0.45\\
    \FMTL & 97.0$\pm$0.05& 97.0$\pm$0.06& 97.4$\pm$0.09& 81.5$\pm$0.24 & 81.5$\pm$0.27 & 89.4$\pm$0.41 & 83.3$\pm$0.20 & 83.2$\pm$0.22 & 90.1$\pm$0.34 \\
    \FedFOMO$^\star$ & 
80.5$\pm$0.75 & 80.4$\pm$0.78 & 95.9$\pm$0.67 & 34.5$\pm$1.57 & 34.3$\pm$1.59 & 90.0$\pm$0.77 & 70.1$\pm$0.56 & 69.9$\pm$0.55 & 89.6$\pm$0.70 \\
    \FedRep$^\star$$\mathsection$ & 95.0$\pm$0.08 & 95.1$\pm$0.11 & 97.5$\pm$0.05 & 79.5$\pm$0.30 & 80.1$\pm$0.31 & 91.8$\pm$0.29 & 80.6$\pm$0.28 & 80.5$\pm$0.34 & 90.5$\pm$0.35 \\
    \midrule
    Local only & - & - & 64.2$\pm$0.68 & - & - & 85.9$\pm$0.69 & - & - & 85.0$\pm$0.80 \\
    \midrule  
    \textbf{\FedRoD (linear)} & \textbf{97.3}$\pm$0.10 & \textbf{97.3}$\pm$0.09 & \textbf{97.5}$\pm$0.09 & \textbf{83.9}$\pm$0.20 & \textbf{83.9}$\pm$0.21 & \textbf{92.7}$\pm$0.24 & \textbf{86.3}$\pm$0.16 & \textbf{86.3}$\pm$0.18 & \textbf{94.5}$\pm$0.20\\
    \textbf{\FedRoD (hyper)} & \textbf{97.3}$\pm$0.10 & \textbf{97.3}$\pm$0.11 & \textbf{97.5}$\pm$0.08 & \textbf{83.9}$\pm$0.18 & \textbf{83.9}$\pm$0.18 & \textbf{92.9}$\pm$0.26 & \textbf{86.3}$\pm$0.17 & \textbf{86.3}$\pm$0.18 & \textbf{94.8}$\pm$0.19\\
    \quad\quad + \FedDyn & \textbf{97.4}$\pm$0.08 & \textbf{97.4}$\pm$0.11 & \textbf{97.5}$\pm$0.11 & \textbf{85.9}$\pm$0.22 & \textbf{85.7}$\pm$0.22 & \textbf{95.3}$\pm$0.36 & \textbf{87.5}$\pm$0.26 & \textbf{87.5}$\pm$0.26 & \textbf{94.6}$\pm$0.35 \\
    \bottomrule
	\end{tabular}
	\label{sup-tbl:main_var1}
\end{table*}

\begin{table*}[t!] 
    \scriptsize
	\centering
	\caption{\small CIFAR-10 results in G-FL accuracy and P-FL accuracy ($\%$). $\star$: methods with no G-FL models and we combine their P-FL models. $\mathsection$: official implementation.} 
	\vspace{-5pt}
	\setlength{\tabcolsep}{3pt}
	\renewcommand{\arraystretch}{0.5}
	\begin{tabular}{l|ccc|ccc|ccc|ccc|ccc|ccc|}
	\toprule {Non-IID} & \multicolumn{3}{|c|}{Dir(0.1)
} & \multicolumn{3}{c}{Dir(0.3)} & \\
    \midrule
    {Test Set} & G-FL & \multicolumn{2}{|c|}{P-FL} & G-FL & \multicolumn{2}{|c|}{P-FL}  \\
    \midrule
    
    {Method / Model} &
    \multicolumn{1}{c|}{GM} & GM & PM & \multicolumn{1}{c|}{GM} & GM & PM \\
    \midrule
    \FedAvg& 57.6$\pm$0.43 & 57.1$\pm$0.42 & 90.5$\pm$0.48 & 68.6$\pm$0.38 & 69.4$\pm$0.41 & 85.1$\pm$0.45 \\ 
    \FedProx& 58.7$\pm$0.21 & 58.9$\pm$0.45 & 89.7$\pm$0.48 & 69.9$\pm$0.39 & 69.8$\pm$0.39 & 84.7$\pm$0.42 \\
    \SCAFFOLD & 61.2$\pm$0.56 & 60.8$\pm$0.59 & 90.1$\pm$0.65 & 71.1$\pm$0.61 & 71.5$\pm$0.60 & 84.8$\pm$0.67 \\
    \FedDyn$\mathsection$& 63.4$\pm$0.40 & 63.9$\pm$0.38 & 92.4$\pm$0.45 & 72.5$\pm$0.37 & 73.2$\pm$0.39 & 85.4$\pm$0.44 \\
    \midrule
    \MOCHA$^\star$ & 12.1$\pm$3.55 & 12.7$\pm$3.78 & 90.6$\pm$0.98 & 13.5$\pm$1.89 & 13.7$\pm$1.93 & 80.2$\pm$1.01 \\
    \LGFedAvg$^\star\mathsection$ & 29.5$\pm$1.46 & 28.8$\pm$1.46 & 90.8$\pm$0.61 & 46.7$\pm$0.45 & 46.2$\pm$0.47 & 82.4$\pm$0.65 \\
    \FedPer$^\star$& 50.4$\pm$0.47 &50.2$\pm$0.48 &89.9$\pm$0.50 & 64.4$\pm$0.44 &64.5$\pm$0.46 &84.9$\pm$0.55 \\
    \PerFedAvg& 60.7$\pm$0.77 & - & 82.7$\pm$1.41 & 70.5$\pm$0.81& - & 80.7$\pm$1.23 \\
    \pFedMe$\mathsection$ & 50.6$\pm$0.56 & 50.7$\pm$0.58 & 76.6$\pm$0.60 & 62.1$\pm$0.60 & 61.7$\pm$0.57 & 70.5$\pm$0.66 \\
    \FMTL & 58.1$\pm$0.49 & 58.3$\pm$0.47 & 86.8$\pm$0.61 & 69.7$\pm$0.44 & 69.8$\pm$0.46 & 81.5$\pm$0.59\\
    \FedFOMO$^\star$ & 30.5$\pm$1.72 & 31.2$\pm$1.74 & 90.5$\pm$0.85 & 45.3$\pm$1.69 & 45.1$\pm$1.66 & 83.4$\pm$0.81\\
    \FedRep$^\star$$\mathsection$  & 56.6$\pm$0.34 & 56.2$\pm$0.35 & 91.0$\pm$0.50 & 67.7$\pm$0.41 & 67.5$\pm$0.33 & 85.2$\pm$0.45 \\
    \midrule
    Local only & - & - & 87.4$\pm$0.69 & - & - & 75.7$\pm$0.78\\
    \midrule
    \textbf{\FedRoD (linear)}  & \textbf{68.5}$\pm$0.35 & \textbf{68.5}$\pm$0.35 & \textbf{92.7}$\pm$0.54 & \textbf{76.9}$\pm$0.37 & \textbf{76.8}$\pm$0.37 & \textbf{86.4}$\pm$0.49 \\
    \textbf{\FedRoD (hyper)}  & \textbf{68.5}$\pm$0.38 & \textbf{68.5}$\pm$0.39 & \textbf{92.5}$\pm$0.55 & \textbf{76.9}$\pm$0.34 & \textbf{76.8}$\pm$0.35 & \textbf{86.8}$\pm$0.55  
\\
    \quad\quad + \FedDyn & \textbf{68.2}$\pm$0.42 & \textbf{68.2}$\pm$0.44 & \textbf{92.7}$\pm$0.57 & \textbf{74.6}$\pm$0.43 & \textbf{74.6}$\pm$0.43 & \textbf{85.6}$\pm$0.58 \\
    \bottomrule
	\end{tabular}
	\label{sup-tbl:main_var2}
\end{table*}
  
\begin{table*}[t!] 
    \scriptsize
	\centering
	\caption{\small CIFAR-100 results in G-FL accuracy and P-FL accuracy ($\%$). $\star$: methods with no G-FL models and we combine their P-FL models. $\mathsection$: official implementation.}    
	\vspace{-5pt}
	\setlength{\tabcolsep}{3pt}
	\renewcommand{\arraystretch}{0.5}
	\begin{tabular}{l|ccc|ccc|ccc|ccc|ccc|ccc}
	\toprule {Non-IID} & \multicolumn{3}{|c|}{Dir(0.1)
} & \multicolumn{3}{c}{Dir(0.3)}\\
    \midrule
    {Test Set} & G-FL & \multicolumn{2}{|c|}{P-FL} & G-FL & \multicolumn{2}{|c|}{P-FL}  \\
    \midrule
    
    {Method / Model} &
    \multicolumn{1}{c|}{GM} & GM & PM & \multicolumn{1}{c|}{GM} & GM & PM \\
    \midrule
    \FedAvg& 41.8$\pm$0.67 &	41.6$\pm$0.71 &	70.2$\pm$0.66 &	46.4$\pm$0.44&	46.2$\pm$0.41&	61.7$\pm$0.40\\ 
    \FedProx& 41.7$\pm$0.51 & 41.6$\pm$0.54 & 70.4$\pm$0.60 & 46.5$\pm$0.48 & 46.4$\pm$0.41 & 61.5$\pm$0.50\\
    \SCAFFOLD &  42.3$\pm$0.73 & 42.1$\pm$0.77 & 70.4$\pm$0.69 & 46.5$\pm$0.68 & 46.5$\pm$0.65 & 61.7$\pm$0.65\\
    \FedDyn$\mathsection$&43.0$\pm$0.39 & 43.0$\pm$0.47 & 72.0$\pm$0.38 & 47.5$\pm$0.41 & 47.4$\pm$0.44 & 62.5$\pm$0.35\\
    \midrule
    \MOCHA$^\star$ & 9.5$\pm$6.55 & 9.3$\pm$5.98 & 60.7$\pm$1.45 & 10.8$\pm$8.71 & 10.7$\pm$6.78 & 49.9$\pm$2.33\\
    \LGFedAvg$^\star\mathsection$ & 23.5$\pm$2.50 & 23.4$\pm$2.14 & 66.7$\pm$1.00 & 34.5$\pm$2.56 & 33.9$\pm$3.01 & 55.4$\pm$1.11 \\
    \FedPer$^\star$& 37.6$\pm$0.65 &	37.6$\pm$0.63 &	71.0$\pm$0.55 &	40.3$\pm$0.51 &	40.1$\pm$0.53 &	62.5$\pm$0.55\\
    \PerFedAvg&  39.0$\pm$0.89 & - & 66.6$\pm$1.12 & 44.5$\pm$0.79 & - & 58.9$\pm$1.30\\
    \pFedMe$\mathsection$ & 38.6$\pm$0.67 & 38.5$\pm$0.65 & 63.0$\pm$0.80 & 41.4$\pm$0.71 & 41.1$\pm$0.68 & 53.4$\pm$0.70\\
    \FMTL & 41.7$\pm$0.56 & 41.8$\pm$0.54 & 68.5$\pm$0.71 & 46.4$\pm$0.45 & 46.4$\pm$0.46 & 58.8$\pm$0.38 &\\
    \FedFOMO$^\star$ & 35.4$\pm$2.00 & 35.3$\pm$1.87 & 68.9$\pm$0.98 & 39.6$\pm$1.89 & 39.3$\pm$1.74 & 58.4$\pm$1.15\\
    \FedRep$^\star$$\mathsection$  & 40.7$\pm$0.51 & 40.7$\pm$0.55 & 71.5$\pm$0.49 & 46.0$\pm$0.37 & 46.0$\pm$0.40 & 62.1$\pm$0.43\\
    \midrule
    Local only & - & - & 40.0$\pm$1.03 & - & - & 32.5$\pm$0.99\\
    \midrule
    \textbf{\FedRoD (linear)}  &  \textbf{45.9}$\pm$0.44 &\textbf{45.8}$\pm$0.41 &	\textbf{72.2}$\pm$0.51 &	\textbf{48.5}$\pm$0.39 &	\textbf{48.5}$\pm$0.38 &	\textbf{62.3}$\pm$0.40\\
    \textbf{\FedRoD (hyper)}  &  \textbf{45.9}$\pm$0.41 &	\textbf{45.8}$\pm$0.39 &	\textbf{72.3}$\pm$0.48 &	\textbf{48.5}$\pm$0.42 &	\textbf{48.5}$\pm$0.45 &	\textbf{62.5}$\pm$0.52 
\\
    \quad\quad + \FedDyn  & 
    \textbf{46.2}$\pm$0.50 &	\textbf{46.2}$\pm$0.51 &	\textbf{72.5}$\pm$0.55 &	\textbf{48.4}$\pm$0.49 &	\textbf{48.4}$\pm$0.47 &	\textbf{62.5}$\pm$0.52\\
    \bottomrule
	\end{tabular}
	\label{sup-tbl:main_var3}
\end{table*}

\clearpage
\section{Additional Experiments and Analyses}
\label{suppl-sec:exp_r}
Here we provide additional experiments and analyses omitted in the main paper. We validate our claims in the main paper and the designs of our proposed \FedRoD via the following experiments:
\begin{itemize}
    \item \autoref{sup_subs_local_strong}: personalized models emerge from local training of generic federated learning (cf. \autoref{ss_gfl_good_pfl}, \autoref{ss_strong_reg}, and \autoref{ss_exp_main} in the main paper).
    \item \autoref{sup_subs_local_meta}: balanced risk minimization (BRM) improves generic-FL performance (cf. \autoref{ss_exp_main} in the main paper).
    \item \autoref{sup_ssec_roles}: the roles of \FedRoD's generic and personalized heads (cf. \autoref{ss_decouple} in the main paper).
    \item \autoref{sup_subs_per_hype}: personalization with hypernetworks (cf. \autoref{ss_exp_main} in the main paper).
    \item \autoref{sup_subs_imb}: robustness to class-imbalanced global data.
    \item \autoref{sup-ss-compatibility}: compatibility of \FedRoD with other G-FL algorithms (cf. \autoref{ss_exp_main} in the main paper).
    \item \autoref{sup-sec-personalized}: comparison to personalized FL algorithms (cf. \autoref{ss_exp_main} in the main paper).
    \item \autoref{sup_subs_arch}: ablation studies and discussions on \FedRoD (cf. \autoref{ss_exp_main} in the main paper).
\end{itemize}

\subsection{Personalized models emerge from local training of generic federated learning}\label{sup_subs_local_strong}

\begin{figure}[t]
    \centering
    \minipage{0.45\columnwidth}
    \centering
    \mbox{Empirical risk}
    \includegraphics[width=1.0\linewidth]{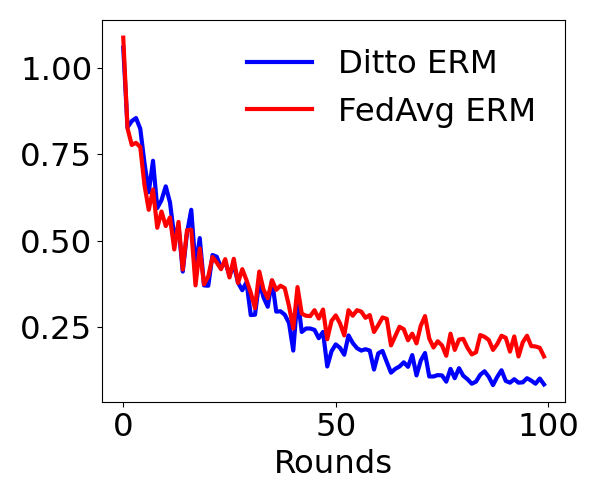} 
    \endminipage
    \minipage{0.45\columnwidth}
    \centering
    \mbox{Regularization}
    \includegraphics[width=1.0\linewidth]{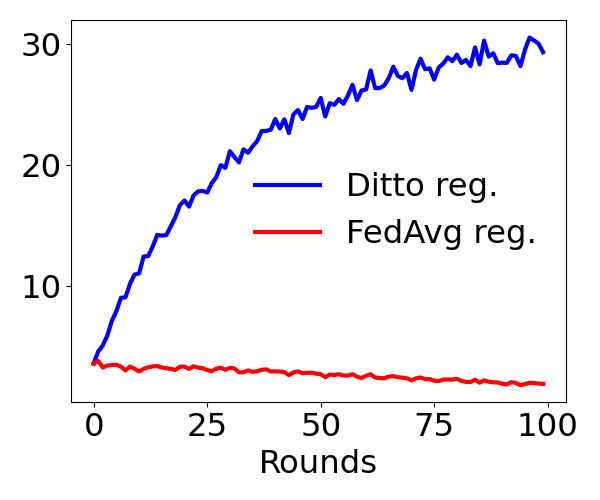}
    \endminipage
    \hspace{0.05cm}
    \caption{\small Comparison of the empirical risk and regularization between personalized models of \FMTL and local models of \FedAvg. The dataset is CIFAR-10, with Dir(0.3).}
    \label{fig:Ditto_grad}
\end{figure}

As mentioned in \autoref{s_GFLPFL} in the main paper, personalized FL algorithms usually impose an extra regularizer (cf. \autoref{eq:P-obj} and \autoref{eq_pfl_local} of the main paper) during local training, but do not re-initialize the local models by the global models at every round. In contrast, generic FL algorithms like \FedAvg do not impose extra regularization but re-initialize the local models at every round. Here in \autoref{fig:Ditto_grad}, we monitor the two loss terms, $\sum_m \frac{|\sD_m|}{|\sD|}\sL_m(\vw_m)$ and $\sum_m \frac{|\sD_m|}{|\sD|}\|\vw_m - \bar{\vw}\|_2^2$ (cf. \autoref{eq:P-obj} and \autoref{eq_pfl_local} of the main paper), for \FedAvg and a state-of-the-art personalized FL algorithm Ditto~\citep{li2020federated-FMTL} at the end of each local training round. (Ditto does include the $L_2$ regularizer in training the personalized models.)
Ditto achieves a lower empirical risk (\ie, the first term), likely due to the fact that it does not perform re-initialization. Surprisingly, \FedAvg achieves a much smaller regularization term (\ie, the second term) than Ditto, even if it does not impose such a regularizer in training. We attribute this to the strong effect of regularization by re-initialization: as mentioned in \autoref{ss_strong_reg} of the main paper, re-initialization is equivalent to setting the regularization coefficient $\lambda$ as infinity. We note that, the reason that the regularization term of Ditto increases along the communication rounds is because ever time the global model $\bar{\vw}$ is updated, it moves sharply away from the local model $\vw_m$. Thus, even if the regularization term is added into local training, it cannot be effectively optimized.
This analysis suggests that the local models of generic FL algorithms are more regularized than the personalized models of personalized FL algorithms. The local models of generic FL algorithms are thus strong candidates to be evaluated in the personalized FL setting.

\subsection{Balanced risk minimization (BRM) improves generic-FL performance}\label{sup_subs_local_meta}

\begin{figure}[t]
    \centering
    \minipage{0.5\columnwidth}
    \centering
    \mbox{\footnotesize \quad FMNIST}
    \includegraphics[width=1\linewidth]{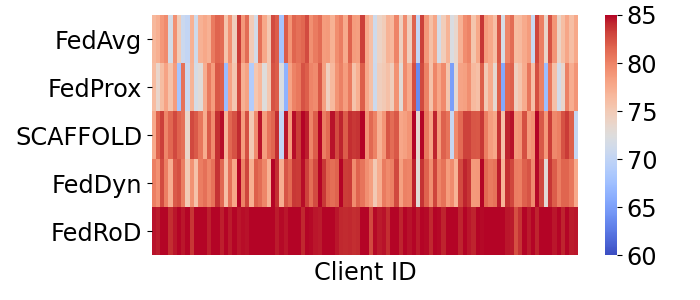} 
    \endminipage \hfill
    \minipage{0.5\columnwidth}
    \centering
    \mbox{\footnotesize \quad CIFAR-10}
    \includegraphics[width=1\linewidth]{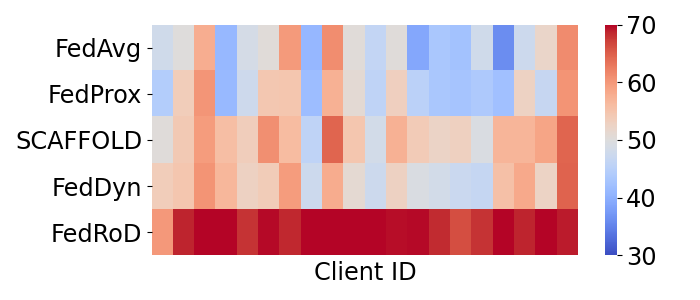}
    \endminipage
    \caption{\small The G-FL accuracy by the local models $\vw_m$ of different generic methods. There are $100/20$ clients for FMNIST/CIFAR-10, respectively. Both datasets use Dir(0.3).}
    \label{sup_fig:local-global}
\end{figure}

\begin{figure}[t]
    \centering
    \minipage{0.45\columnwidth}
    \centering
    \mbox{FMNIST}
    \includegraphics[width=1.0\linewidth]{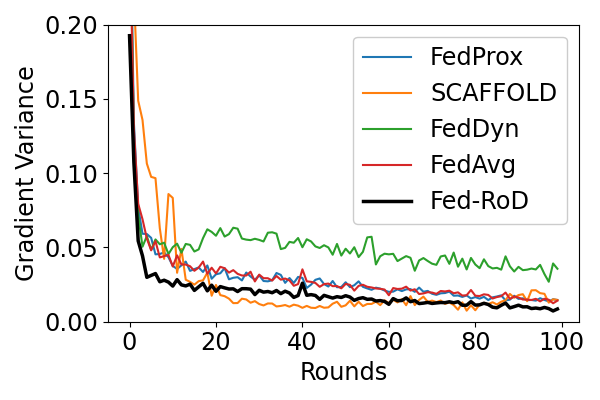} 
    \endminipage
    \minipage{0.45\columnwidth}
    \centering
    \mbox{CIFAR-10}
    \includegraphics[width=1.0\linewidth]{figures/grad_cifar10.png}
    \endminipage
    \hspace{0.05cm}
    \caption{\small Variances of local model updates w.r.t. the global model. For both datasets, we use Dir(0.3).}
    \label{sup_fig:grad}
\end{figure}

To understand why \FedRoD outperforms other generic methods in the G-FL accuracy, we visualize each \emph{local} model $\vw_m$'s G-FL accuracy after local training in~\autoref{sup_fig:local-global}  (both datasets with Dir(0.3)). Methods rely on ERM suffer as their local models tend to diverge. 
\autoref{sup_fig:grad} further shows that the variances of local weight update $\Delta \vw_m = \vw_m - \bar{\vw}$ across clients are smaller for \FedRoD, which result from a more consistent local training objective. 

\begin{figure*}[t]
    \centering
    {\includegraphics[width=0.6\textwidth]{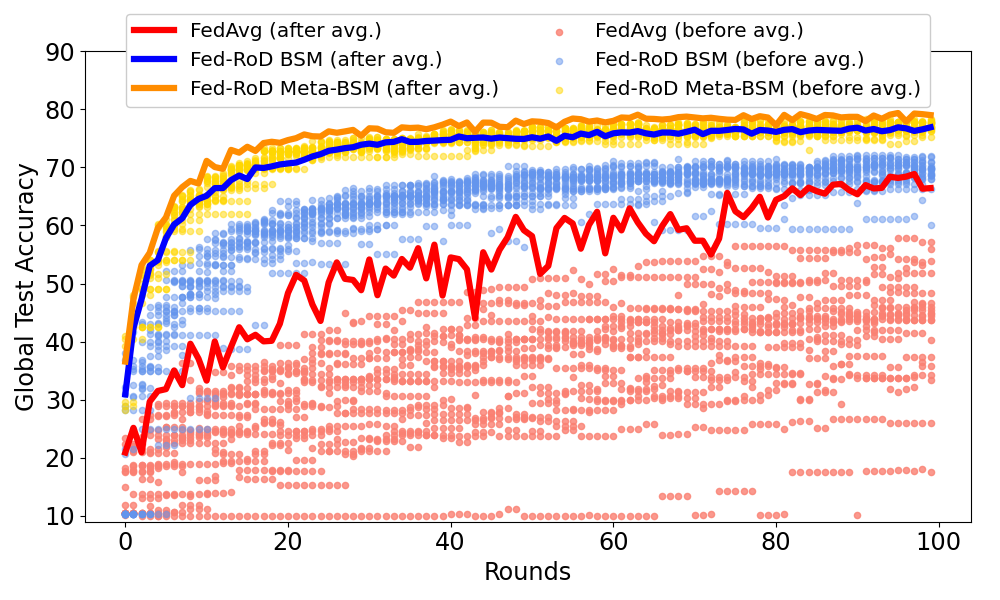}}%
    \caption{\small Training curves of different FL algorithms. We show the G-FL accuracy along the training process, using models before (\ie, local models) and after global aggregation. The dataset is CIFAR-10 Dir(0.3).}
    \label{sup_fig:global_curve_meta}    
\end{figure*}

In \autoref{sup_fig:global_curve_meta}, we further compare the G-FL accuracy among \FedAvg, \FedRoD with the original BSM loss, and \FedRoD with the Meta-BSM loss introduced in~\autoref{sup_subsec_meta} along the training process (\ie, training curve).
The local models of \FedAvg tend to diverge from each other due to the non-IID issue, resulting in high variances and low accuracy of G-FL. The global aggregation does improve the G-FL accuracy, validating its importance in federated learning. 
The local training in \FedRoD (BSM) not only leads to a better global model, but also has smaller variances and higher accuracy for the local models (as their objectives are more aligned).
With the help of meta dataset and meta-learning, \FedRoD (Meta-BSM) yields even better G-FL performance for both global models and local models, and has much smaller variances among local models' performance, demonstrating the superiority of using meta-learning to learn a balanced objective. 

\begin{figure}[h!]
    \centering
    {\includegraphics[width=0.8\columnwidth]{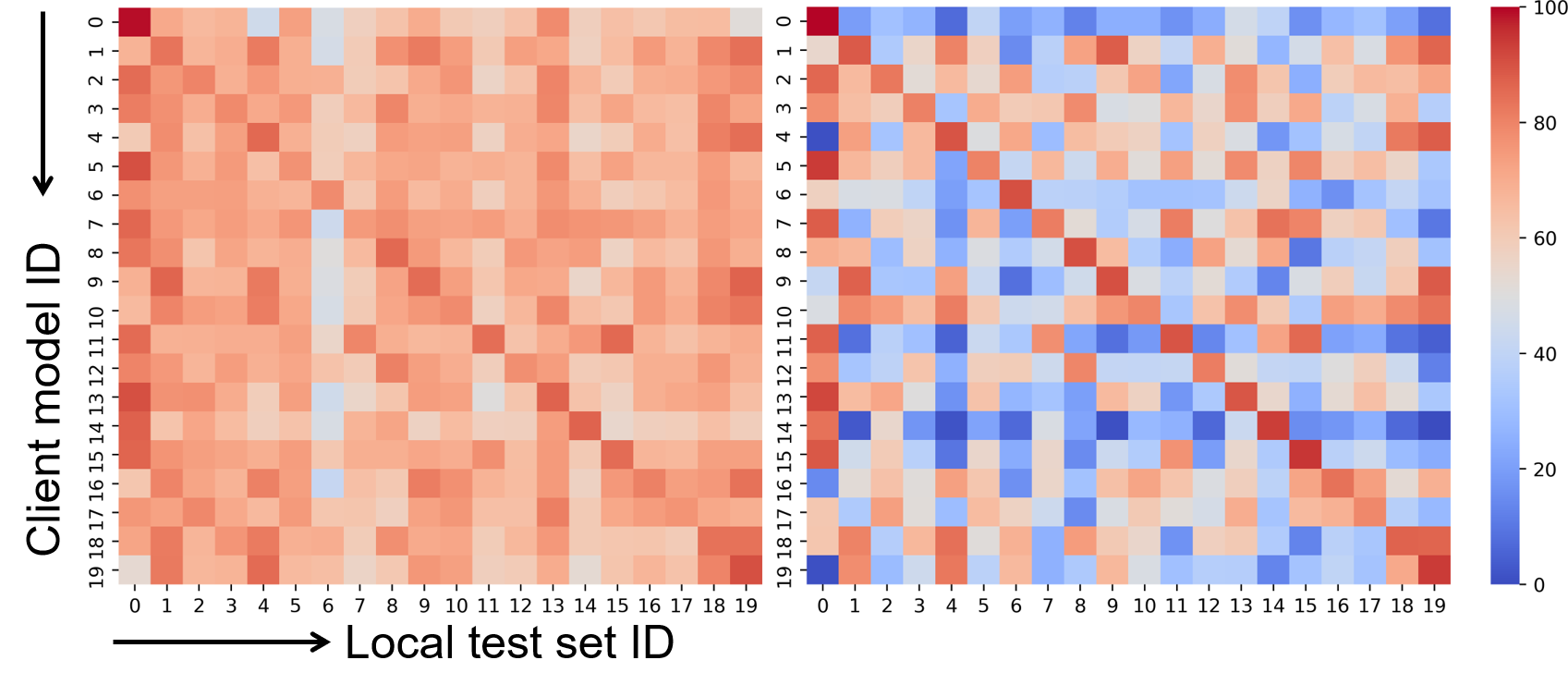}}%
    \caption{\small P-FL accuracy of G-head (left) and G-head + P-head (right) using the local models of \FedRoD, evaluated on each client's test data. Here we use CIFAR-10 Dir(0.3) with 20 clients.}
    \label{sup-fig:client_acc_confusion}
\end{figure} 

\begin{figure}[t]
    \centering
    {\includegraphics[width=.6\columnwidth]{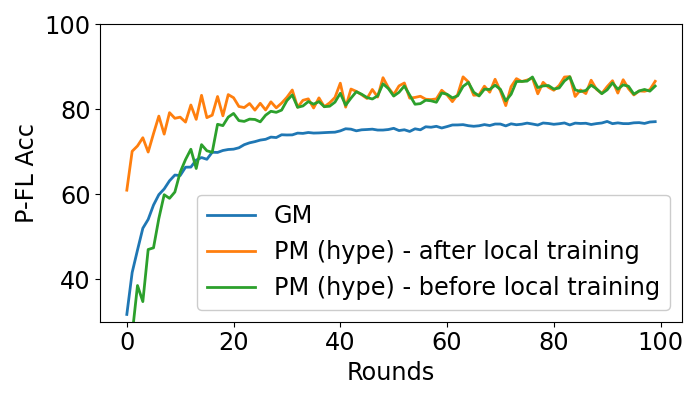}}%
    \caption{\small P-FL accuracy of hypernetwork before and after local training.}
    \label{fig:hype_curve}  
\end{figure}

\subsection{The roles of \FedRoD's generic and personalized heads} \label{sup_ssec_roles} 

To demonstrate that \FedRoD's two heads learn something different, we plot in \autoref{sup-fig:client_acc_confusion} every local model's generic prediction and personalized prediction on its and other clients' data (\ie, P-FL accuracy). The generic head performs well in general for every client's test data. The personalized head could further improve for its own data (diagonal), but degrade for others' data.


\subsection{Personalization with hypernetworks}
\label{sup_subs_per_hype}
\FedRoD (hyper) learns the personalized head with hypernetworks as introduced in~\autoref{sup_subsec_hype}. The goal is to learn a hypernetwork such that it can directly generate a personalization prediction head given client's class distribution, without further local training. \autoref{fig:hype_curve} shows the training (convergence) curves on CIFAR-10 Dir(0.3). The hypernetwork (globally aggregated, before further local training) can converge to be on par with that after local training. In the main paper (cf. \autoref{fig:new_clients}), we also show that it servers as a strong starting point for future clients --- it can generate personalized models simply with future clients' class distributions. That is, the clients may not have labeled data, but provide the hypernetwork with their preference/prior knowledge.
It can also be used as the warm-start model for further local training when labeled data are available at the clients.

\autoref{tbl:all_local_bf_aft} provides the P-FL results for the new 50 clients studied in \autoref{ss_exp_main} and \autoref{fig:new_clients} in the main paper. Except for \FedRoD (hyper), the accuracy before local training is obtained by the global model. The best personalized model after local training is selected for each client using a validation set.
\FedRoD (hyper) notably outperforms other methods before or after local training.


\definecolor{Gray}{gray}{0.9}    
\begin{table*}[t] 
    \footnotesize
	\centering
	\caption{\small Class-imbalanced global training distribution.  $^\star$: methods with no global models and we combine their P-FL models. Gray rows: meta-learning with 100 labeled server data.} 
	\setlength{\tabcolsep}{1.5pt}
	\renewcommand{\arraystretch}{0.5}
	\begin{tabular}{l|ccc|ccc|ccc|ccc|ccc|ccc|ccc|ccc}
	\toprule {Dataset} & \multicolumn{6}{c|}{FMNIST} & \multicolumn{6}{c}{CIFAR-10}\\
	\midrule {Non-IID / Imbalance Ratio} & \multicolumn{3}{c|}{Dir(0.3), IM10
} & \multicolumn{3}{c}{Dir(0.3), IM100} & \multicolumn{3}{|c|}{Dir(0.6), IM10
} & \multicolumn{3}{c}{Dir(0.6), IM100}\\
    \midrule
    {Test Set} & G-FL & \multicolumn{2}{|c|}{P-FL} & G-FL & \multicolumn{2}{|c|}{P-FL} & G-FL & \multicolumn{2}{|c|}{P-FL} & G-FL & \multicolumn{2}{|c}{P-FL}\\
    \midrule
    
    {Method / Model} & \multicolumn{1}{c|}{GM} & GM & PM & \multicolumn{1}{c|}{GM} & GM & PM & \multicolumn{1}{c|}{GM} & GM & PM & \multicolumn{1}{c|}{GM} & GM & PM\\
    \midrule
    
    \FedAvg~\citep{mcmahan2017communication}&  80.2 & 80.2 & 85.1 & 71.6 &71.5 & 86.9 & 50.9 & 50.2 & 76.5 & 40.1 & 40.0 & 78.2\\ 
    \FedProx~\citep{li2020federated}& 81.0 & 81.0 & 82.3 & 70.4 & 70.2 & 87.0 & 58.6 & 58.7 & 76.5 & 37.6 & 38.0 &  76.6 \\
    \SCAFFOLD~\citep{karimireddy2020scaffold}& 81.1 & 81.1 & 82.2 & 72.0 & 71.8 & 86.9 & 58.7 & 58.7 & 76.6 & 38.4 & 38.4 & 77.6\\
    \FedDyn~\citep{feddyn}& 83.3 & 83.2 &86.4 & 77.2 & 77.1 & 87.5 & 62.5 & 62.3 & 80.4 & 46.6 & 46.5 & 80.9\\
    \midrule
    \MOCHA~\citep{smith2017federated}$^\star$ & 45.3 & 45.4 & 75.6 & 47.7 & 47.5 & 77.5 & 17.9 & 18.2 & 63.4 & 14.4 & 14.9 & 65.9 \\
    \LGFedAvg~\citep{liang2020think}$^\star$ & 62.8 & 62.4 & 82.4 & 74.0 & 74.1 & 83.2 & 31.5 & 31.5 & 62.8 & 24.9 & 24.8 & 66.3 \\
    \PerFedAvg~\citep{fallah2020personalized} & 80.1 & - & \multicolumn{1}{c|}{82.5} & 72.0 & - & \multicolumn{1}{c|}{78.5} & 46.3 & - & \multicolumn{1}{c|}{77.2} & 31.7 & - & \multicolumn{1}{c|}{74.3}\\
    \pFedMe~\citep{dinh2020personalized} & 78.9 & 78.9 & 81.6 & 69.3 & 69.2 & 71.6 & 46.2 & 46.2 & 54.2 & 31.7 & 31.8 & 50.6\\
    \FMTL~\citep{li2020federated-FMTL} & 81.0 & 81.0 & 83.7 & 71.8 & 71.6 & 86.5 & 51.0 & 50.9 & 73.1 & 40.3& 40.2 & 75.4\\
    \FedFOMO~\citep{fedfomo}$^\star$ & 65.5 & 65.2 & 89.5 & 64.5 & 64.4 & \textbf{90.1} & 42.7 & 42.6 & 76.6 & 23.6 & 23.8 &76.7\\
    \midrule
    Local only & - & - & 76.1 & - & - & 79.8 & - & - & 72.1 & - & - & 74.5\\

    \midrule
    \FedRoD (BSM) & 81.3 & 81.3 & 89.5 & 76.8 & 76.8 &89.8 & 63.3 & 63.3 & \textbf{80.1} & 48.3 & 48.3 & \textbf{81.1} \\
    \FedRoD (BSM) + \FedDyn & \textbf{84.5} & \textbf{84.5} & \textbf{90.0} & \textbf{80.0} & \textbf{79.9} & 88.6 & \textbf{65.5} & \textbf{65.5} & 79.7 & \textbf{49.6} & \textbf{49.6} & 81.0\\
    \midrule
    \midrule
    \rowcolor{Gray}
    \FedAvg + Meta~\citep{zhao2018federated} & 80.5 & 80.5 & 85.1 & 71.8 & 71.8 & 86.9 & 51.5 & 51.3 & 77.4 & 40.4 & 40.2 & \textbf{84.0}\\
    \rowcolor{Gray}
    \FedRoD (Meta-BSM) & \textbf{86.5} & \textbf{86.5} & \textbf{90.2} & \textbf{82.5}& \textbf{82.5} & \textbf{90.9} & \textbf{72.3} & \textbf{72.1} & \textbf{82.1} & \textbf{61.9} & \textbf{60.9} & 83.5\\
    \bottomrule
	\end{tabular}
	\label{tbl:main_imb}
\end{table*}

\begin{table}[t] 
    \footnotesize
	\centering
	\caption{\small G-FL accuracy on class-imbalanced test data. Here we use CIFAR-10 Dir(0.3).} 
	\setlength{\tabcolsep}{4pt}
	\renewcommand{\arraystretch}{0.5}
	\begin{tabular}{l|cc}
	\toprule
	Method  &  IM10 & IM100 \\
    \midrule
    \FedAvg~\citep{mcmahan2017communication} & 61.8 & 73.0\\
    \FedProx~\citep{li2020federated} & 63.1 & 72.9 \\
    \SCAFFOLD~\citep{karimireddy2020scaffold} & 65.4 & 70.4\\
    \FedDyn~\citep{feddyn} & 68.6 & 73.3 \\
    \FedRoD & 71.9 & 76.0\\
    \bottomrule
	\end{tabular}
	\label{tbl:imb_test}
\end{table}

\subsection{Class-imbalanced global distributions}
\label{sup_subs_imb}
In the real world, data frequency naturally follows a long-tailed distribution, rather than a class-balanced one. Since the server has no knowledge and control about the whole collection of the clients' data, the clients data may collectively be class-imbalanced. This adds an additional challenge for the server to learn a fair and class-balanced model. We follow the setup in~\citep{cao2019learning} to transform FMNIST and CIFAR-10 training sets into class-imbalanced versions, in which the sample sizes per class follow an exponential decay. The imbalanced ratio (IM) is controlled as the ratio between sample sizes of the most frequent and least frequent classes. Here we consider IM$=10$ and IM$=100$. The generic test set remains class-balanced. 

\autoref{tbl:main_imb} shows that \FedRoD remains robust on both generic accuracy and client accuracy consistently. We see that \FedDyn also performs well, especially on FMNIST of which the setup has more clients ($100$) but a lower participation rate ($20\%$). 
By combining \FedDyn with \FedRoD, we achieve further improvements.

Essentially, the generic FL methods (except for \FedRoD) are optimizing toward the overall class-imbalanced distribution rather than the class-balanced distribution. In~\autoref{tbl:imb_test}, we further examine the G-FL accuracy on a class-imbalanced test set whose class distribution is the same as the global training set. \FedRoD still outperforms other methods, demonstrating that \FedRoD learns a robust, generic, and strong model. 


\definecolor{Gray}{gray}{0.9}
\begin{table*}[t!] 
    \footnotesize
	\centering
	\caption{\small Main results in G-FL accuracy and P-FL accuracy ($\%$), following \autoref{tbl:main_bal} of the main paper. \FedRoD is compatible with other generic FL methods.} 
	\setlength{\tabcolsep}{1.5pt}
	\renewcommand{\arraystretch}{0.35}
	\begin{tabular}{l|ccc|ccc|ccc|ccc|ccc|ccc|ccc|ccc}
	\toprule {Dataset} & \multicolumn{6}{c|}{FMNIST} & \multicolumn{6}{c}{CIFAR-10}\\
	\midrule {Non-IID} & \multicolumn{3}{c|}{Dir(0.1)
} & \multicolumn{3}{c}{Dir(0.3)} & \multicolumn{3}{|c|}{Dir(0.1)
} & \multicolumn{3}{c}{Dir(0.3)}\\
    \midrule
    {Test Set} & G-FL & \multicolumn{2}{|c|}{P-FL} & G-FL & \multicolumn{2}{|c|}{P-FL} & G-FL & \multicolumn{2}{|c|}{P-FL} & G-FL & \multicolumn{2}{|c}{P-FL}\\
    \midrule
    
    {Method / Model} & \multicolumn{1}{c|}{GM} & GM & PM & \multicolumn{1}{c|}{GM} & GM & PM & \multicolumn{1}{c|}{GM} & GM & PM & \multicolumn{1}{c|}{GM} & GM & PM\\
    \midrule
    \FedAvg~\citep{mcmahan2017communication}&  81.1 & 81.0 & 91.5 & 83.4 &83.2 & 90.5 & 57.6 & 57.1 & 90.5 & 68.6 & 69.4 & 85.1\\ 
    \FedProx~\citep{li2020federated}& 82.2 & 82.3& 91.4 & 84.5 & 84.5 & 89.7 & 58.7 & 58.9 & 89.7 & 69.9 & 69.8 & 84.7\\
    \SCAFFOLD~\citep{karimireddy2020scaffold}& 83.1 & 83.0 & 89.0 & 85.1 & 85.0 & 90.4 & 61.2 & 60.8 & 90.1 & 71.1 & 71.5 & 84.8 \\
    \FedDyn~\citep{feddyn}$\mathsection$& 83.2 & 83.2 &90.7 & 86.1 & 86.1 & 91.5 & 63.4 & 63.9 & 92.4 & 72.5 & 73.2 & 85.4\\
    \midrule
    \FedRoD (linear) & 83.9& 83.9 & 92.7 & 86.3 & 86.3 & 94.5 & 68.5 & 68.5 & 92.7 & 76.9 & 76.8 & 86.4\\
    \FedRoD (hyper)& 83.9 & {83.9} & {92.9} & {86.3} & {86.3} & {94.8} & {68.5} & {68.5} & {92.5} & {76.9} & {76.8} & {86.8}\\
    \quad\quad + \FedProx & {83.3} & {83.3} & {93.8} & {85.8} & {85.7} & {92.2} & {70.6} & {70.5} & {92.5} & {74.5} & {74.5} & {85.7} &\\
    \quad\quad + \SCAFFOLD & {84.3} & {84.3} & {94.8} & {88.0} & {88.0} & {94.7} & {72.0} & {71.8} & {92.6} & {77.8} & {77.7} & {86.9} &\\
    \quad\quad + \FedDyn & {85.9} & {85.7} & {95.3} & {87.5} & {87.5} & {94.6} & {68.2} & {68.2} & {92.7} & {74.6} & {74.6} & {85.6} &\\
    \midrule
    \midrule
    \rowcolor{Gray}
    \FedAvg + Meta~\citep{zhao2018federated} & 83.1 & 83.1 & 91.5 & 84.4 & 84.3 & 90.5 & 58.7 & 58.9 & 90.5 & 69.2 & 69.2 & 85.3\\
    \rowcolor{Gray}
    \textbf{\FedRoD (Meta-BSM)} & \textbf{86.4} & \textbf{86.4} & \textbf{94.8} & \textbf{89.1} & \textbf{89.1} & \textbf{94.8} & \textbf{72.5} & \textbf{72.5} & \textbf{92.8} & \textbf{80.1} & \textbf{80.1} & \textbf{86.6}\\
    \bottomrule
	\end{tabular}
	\label{sup_tbl:main_bal}
\end{table*}

\subsection{Compatibility of \FedRoD with other G-FL algorithms}
\label{sup-ss-compatibility}

As mentioned in the main paper, other G-FL algorithms like \FedDyn~\citep{feddyn} can be incorporated into \FedRoD to optimize the generic model (using the balanced risk). We show the results in \autoref{sup_tbl:main_bal}, following \autoref{tbl:main_bal} of the main paper. Combining \FedRoD with \SCAFFOLD~\citep{karimireddy2020scaffold}, \FedDyn~\citep{feddyn}, and \FedProx~\citep{li2020federated} can lead to higher accuracy than each individual algorithm along in several cases.

\begin{table}[] 
    \footnotesize
	\centering
	\caption{\small The P-FL accuracy by the two local models of \FMTL~\citep{li2020federated-FMTL}.} 
	\setlength{\tabcolsep}{4pt}
	\renewcommand{\arraystretch}{0.5}
	\begin{tabular}{l|cc|cc}
	\toprule
	Method  &  \multicolumn{2}{c}{FMNIST} & \multicolumn{2}{|c}{CIFAR-10} \\
	\midrule
	& Dir(0.1) & Dir(0.3) & Dir(0.1) & Dir(0.3) \\ 
    \midrule
    PM & 89.4 & 90.1 & 86.8 & 81.5\\
    LM & 90.8 & 90.6 & 90.8 & 86.2\\
    \bottomrule
	\end{tabular}
	\label{tbl:fmtl}
\end{table}

\begin{table}[] 
    \footnotesize
	\centering
	\caption{\small \FMTL with adversary attacks. We report the averaged personalized accuracy on benign clients.} 
	\setlength{\tabcolsep}{4pt}
	\renewcommand{\arraystretch}{0.75}
	\begin{tabular}{l|ccc}
	\toprule
	Attack  &  PM & LM & GM\\
    \midrule
    None & 94.2 & 94.7 & 91.7\\
    Label poisoning & 93.6 & 54.5 & 84.8\\
    Random updates & 93.2 & 54.5 & 88.7\\
    Model replacement & 63.6 & 49.8 & 42.2\\
    \bottomrule
	\end{tabular}
	\label{tbl:fmtl_adv}
\end{table}

\subsection{Comparison to personalized FL algorithms}
\label{sup-sec-personalized}

From \autoref{tbl:main_bal} in the main paper and \autoref{tbl:main_imb}, the personalized FL algorithms are usually outperformed by local models of generic FL algorithms in terms of the P-FL accuracy (\ie, the PM column). The gap is larger when client data are more IID, especially for P-FL methods whose personalized models do not explicitly rely on weight averaging of other clients' models (\eg, \MOCHA, \LGFedAvg, and \pFedMe). Some P-FL methods can not even outperform local training alone. A similar observation is also reported in \FedFOMO~\citep{fedfomo}. These observations justify the benefits of FL that similar clients can improve each other by aggregating a global model and updating it locally, while the benefits might decay for very dissimilar clients.

To further demonstrate the effect of building a global model and re-initialing the local/personalized models using it (cf. \autoref{s_GFLPFL} in the main paper), we investigate \FMTL~\citep{li2020federated-FMTL}, a state-of-the-art personalized FL algorithm. We found that \FMTL learns \emph{two} local models. One of them is used to build the global model exactly like \FedAvg. The global model is then used to regularize the other local model (cf. \autoref{eq_pfl_local} in the main paper), which is used for personalized prediction. To differentiate these two local models, we call the former the local model (LM), and the latter the personalized model (PM). We note that, the PM model is kept locally and is never re-initialized by the global model.
In \autoref{tbl:fmtl}, we show the P-FL accuracy using the LM and PM models. The LM model trained in the same way as \FedAvg (with re-initialization) surprisingly outperforms the PM model.
  
We further replicate the experiments in~\citep{li2020federated-FMTL} on robustness against adversary attacks in~\autoref{tbl:fmtl_adv}. Besides comparing LM and PM, we also evaluate the global model GM for P-FL accuracy. With out adversarial attacks, the LM model outperforms the PM model. However, with adversarial attacks, the PM model notably outperforms the other two models. We surmise that, when there are adversarial clients, the resulting generic model will carry the adversarial information; re-initializing the local models with it thus would lead to degraded performance.

\begin{table}[] 
    \footnotesize
	\centering
	\caption{\small \FedRoD on CIFAR-10, Dir(0.3).} 
	\setlength{\tabcolsep}{4pt}
	\renewcommand{\arraystretch}{0.5}
	\begin{tabular}{l|ccc}
    \midrule
    {Test Set} & G-FL & \multicolumn{2}{|c}{P-FL} \\
    \midrule
    {Network} & \multicolumn{1}{c|}{GM} & GM & PM \\
    \midrule
    ConvNet~\citep{lecun1998gradient} & 76.9 & 76.8 & 86.8 \\
    VGG11~\citep{Simonyan15} & 82.2 & 82.1 & 88.2 \\
    ResNet8~\citep{he2016deep} & 80.3 & 80.0 & 86.6 \\
    ResNet20~\citep{he2016deep} & 84.0 & 83.5 & 88.5 \\
    \bottomrule
	\end{tabular}
	\label{tbl:ablation_deep}
\end{table}

\subsection{Additional studies and discussions}
\label{sup_subs_arch}

\paragraph{Different network architectures.}
\FedRoD can easily be applied to other modern neural network architectures. In~\autoref{tbl:ablation_deep}, we show that \FedRoD can be used with deeper networks.

\paragraph{\FedRoD is not merely fine-tuning.}
\FedRoD is not merely pre-training the model with BSM and then fine-tuning it with ERM for two reasons. First, for \FedRoD (linear), the P-head is learned dynamically with the updating feature extractor across multiple rounds. Second, for \FedRoD (hyper), the hypernetwork has to undergo the local training and global aggregation iterations over multiple rounds.
In \autoref{tbl:ablation_irm} of the main paper, we report the fine-tuning baseline. 
On CIFAR-10 Dir(0.3), it has $84.5\%$ for P-FL (PM), lower than $86.4\%$ and $86.8\%$ by \FedRoD (linear) and \FedRoD (hyper). {Note that, hypernetworks allow fast adaptation for new clients.}

\paragraph{Comparison to the reported results in other personalized FL papers.} Existing works usually report \FedAvg's personalized performance by evaluating its global model (\ie, the GM column in \autoref{tbl:main_bal} of the main paper). In this paper, we evaluate \FedAvg's \emph{local} model $\vw_m$ (\ie, the PM column in \autoref{tbl:main_bal} of the main paper), which is locally trained for epochs. We see a huge performance gap between these two models. In~\citep{fallah2020personalized}, the authors investigated a baseline ``\FedAvg + update'', which fine-tunes \FedAvg's \emph{global} model $\bar{\vw}$ with only few mini-batches for each client. The resulting personalized models thus capture less personalized information than $\vw_m$ in \FedAvg.
For a fair comparison, we also strengthen \PerFedAvg~\citep{fallah2020personalized} by updating with more epochs.

\paragraph{Effects of local sample size to P-FL performance}
In~\autoref{ss_gfl_good_pfl} and~\autoref{tbl:main_bal}, we show that  local models of generic FL algorithms are strong personalized models. Indeed, the local sample size is an important factor in the P-FL performance. If a client has enough training samples, training its own model (the \emph{local only} baseline) can already be strong without any federated learning. On the other hand, when each client does not have enough samples to train a good model on its own. It will be crucial to have a generic model learned from federated learning as the starting point of personalization. 
 
To confirm our observation when clients have insufficient samples, we further conduct the following experiments. First, we enlarge the number of clients for CIFAR-10 and FMNIST experiments by five times. That is, each client’s data size becomes one-fifth on average. Second, we point out that the experiments on CIFAR-100 in~\autoref{tbl:main_bal} are with 20 clients. CIFAR-100 has the same total number of training images as CIFAR-10 but with 10 times more classes. In other words, the number of images per class is one-tenth. ~\autoref{sup_tbl:more_clients} shows the results: all the experiments are based on Dir(0.3). Even when the \emph{local only} models perform worse in P-FL, the local models of \FedAvg still perform on a par with personalized FL algorithms like \FedPer, and \FedRoD can still achieve the best P-FL accuracy. We attribute the superior personalized performance by \FedAvg and \FedRoD to the implicit regularization discussed in~\autoref{ss_strong_reg}.

We also want to point out that, even if each client has insufficient data, the P-FL performance of \emph{local only} may still have higher accuracy than the GM of \FedAvg on the personalized accuracy, especially when the non-IID condition becomes severe (e.g., Dir (0.1)). When the non-IID condition is severe, it is harder to train a single GM model to perform well in the personalized setting.

\begin{table*}[t!] 
    \footnotesize
	\centering
	\caption{\small Main results in G-FL accuracy and P-FL accuracy ($\%$), following \autoref{tbl:main_bal} of the main paper. \FedRoD is compatible with other generic FL methods.} 
	\setlength{\tabcolsep}{1.5pt}
	\renewcommand{\arraystretch}{0.35}
	\begin{tabular}{l|ccc|ccc|ccc|}
	\toprule {Dataset} & \multicolumn{3}{c|}{FMNIST} & \multicolumn{3}{c}{CIFAR-10} & \multicolumn{3}{c}{CIFAR-100}\\
    \midrule
    {Test Set} & 
    G-FL & \multicolumn{2}{|c|}{P-FL} & 
    G-FL & \multicolumn{2}{|c|}{P-FL} & 
    G-FL & \multicolumn{2}{|c|}{P-FL} \\
    \midrule
    
    {Method / Model} & 
    \multicolumn{1}{c|}{GM} & GM & PM & \multicolumn{1}{c|}{GM} & GM & PM & \multicolumn{1}{c|}{GM} & GM & PM \\

    \midrule
    Local only & - & - & 72.9 & - & - & 76.9 & - & - & 32.5\\
    \midrule
    \FedAvg~\citep{mcmahan2017communication}&  78.1 &	77.9&	85.7&	64.2&	64.0&	77.4&	46.4&	46.2&	61.7\\ 
    \FedPer~\citep{arivazhagan2019federated}&  72.5&	72.4&	85.5&	57.6&	55.9&	78.0&	40.3&	40.1&	62.5\\
    \midrule  
    \FedRoD (hype)& 82.6&	82.6&	90.1&	72.7&	72.7&	82.7&	48.5&	48.5&	62.5\\
    \bottomrule
	\end{tabular}
	\label{sup_tbl:more_clients}
\end{table*}

\end{document}